\documentclass[10pt,twocolumn,letterpaper]{article}
\usepackage{iccv}

\iccvfinalcopy

\usepackage{times}
\usepackage{epsfig}
\usepackage{graphicx}
\usepackage{amsmath}
\usepackage{amssymb}
\usepackage{xcolor}
\usepackage{import}
\usepackage{booktabs}
\usepackage{multirow}
\usepackage{caption}
\usepackage{subcaption}
\usepackage{pifont}
\usepackage{multirow}
\usepackage{xspace}
\usepackage{epstopdf}
\usepackage{float}
\usepackage[export]{adjustbox}
\usepackage{sidecap}

\usepackage[pagebackref=true,breaklinks=true,letterpaper=true,colorlinks,bookmarks=false]{hyperref}

\newcommand{\rot}[2][0]{\rotatebox{#1}{#2}}

\newcommand{\CE}{\texttt{{CE}}\xspace}
\newcommand{\MOCO}{\texttt{{SelfSupCon}}\xspace}
\newcommand{\MIT}{\texttt{{SupCon}}\xspace}
\newcommand{\MITplusMOCO}{\texttt{{SupCon+SelfSupCon}}\xspace}
\newcommand{\CEplusMOCO}{\texttt{{CE+SelfSupCon}}\xspace}
\newcommand{\CEstrong}{\texttt{{CE(strong)}}\xspace}
\newcommand{\CEmoco}{\texttt{{CE(MoCo-aug)}}\xspace}

\graphicspath{{ICCV2021/}}

\def\iccvPaperID{9362} 

\ificcvfinal\fi

\begin{document}

\title{A Broad Study on the Transferability of Visual Representations with \\ Contrastive Learning}

\author{Ashraful Islam\textsuperscript{1}\thanks{This work was done while the author was an intern at IBM.},
Chun-Fu Chen\textsuperscript{2,3},
Rameswar Panda\textsuperscript{2,3},
Leonid Karlinsky\textsuperscript{3},
Richard Radke\textsuperscript{1},
Rogerio Feris\textsuperscript{2,3} 
\\
\textsuperscript{1}{Rensselaer Polytechnic Institute} \quad
\textsuperscript{2}{MIT-IBM Watson AI Lab} \quad \textsuperscript{3}{IBM Research} \\

{\tt\small islama6@rpi.edu, chenrich@us.ibm.com, rpanda@ibm.com}\\ {\tt\small leonidka@il.ibm.com, rjradke@ecse.rpi.edu, rsferis@us.ibm.com}
} 

\maketitle

\ificcvfinal\thispagestyle{empty}\fi

\begin{abstract}

Tremendous progress has been made in visual representation learning, notably with the recent success of self-supervised contrastive learning methods. Supervised contrastive learning has also been shown to outperform its cross-entropy counterparts by leveraging labels for choosing where to contrast. However, there has been little work to explore the transfer capability of contrastive learning to a different domain. In this paper, we conduct a comprehensive study on the transferability of learned representations of different contrastive approaches for linear evaluation, full-network transfer, and few-shot recognition on 12 downstream datasets from different domains, and object detection tasks on MSCOCO and VOC0712. The results show that the contrastive approaches learn representations that are easily transferable to a different downstream task. We further observe that the joint objective of self-supervised contrastive loss with cross-entropy/supervised-contrastive loss leads to better transferability of these models over their supervised counterparts. Our analysis reveals that the representations learned from the contrastive approaches contain more low/mid-level semantics than cross-entropy models, which enables them to quickly adapt to a new task. Our codes and models will be publicly available to facilitate future research on transferability of visual representations. \footnote{\url{https://github.com/asrafulashiq/transfer_broad}}
\vspace{-3mm}

\end{abstract}


\section{Introduction}
\label{sec:intro}

\begin{figure}[tb]
    \centering
    \begin{subfigure}{0.28\textwidth}
    \centering
        \includegraphics[height=2.1cm]{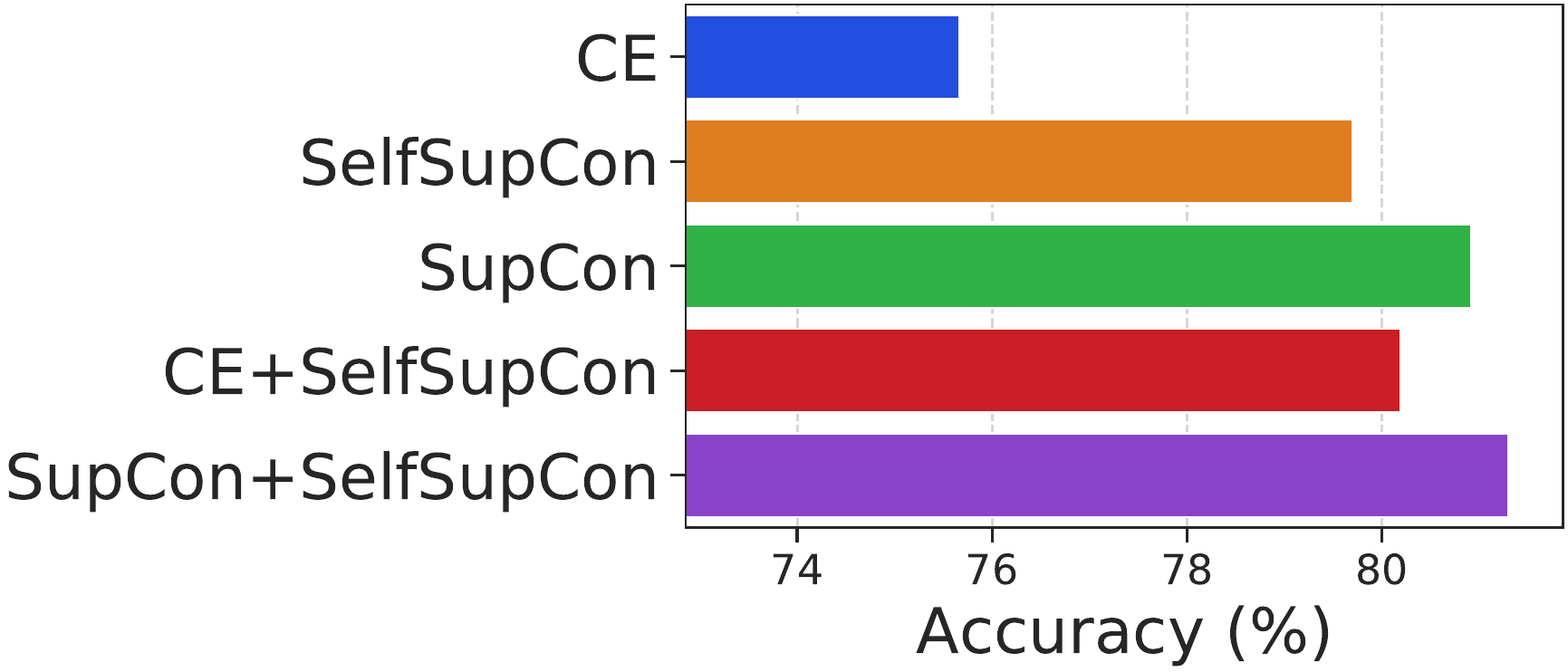}
        \caption{{\small Linear evaluation}}
        \label{fig:teaser1}
    \end{subfigure}
    \begin{subfigure}{0.19\textwidth}
    \centering
        \includegraphics[height=2.1cm]{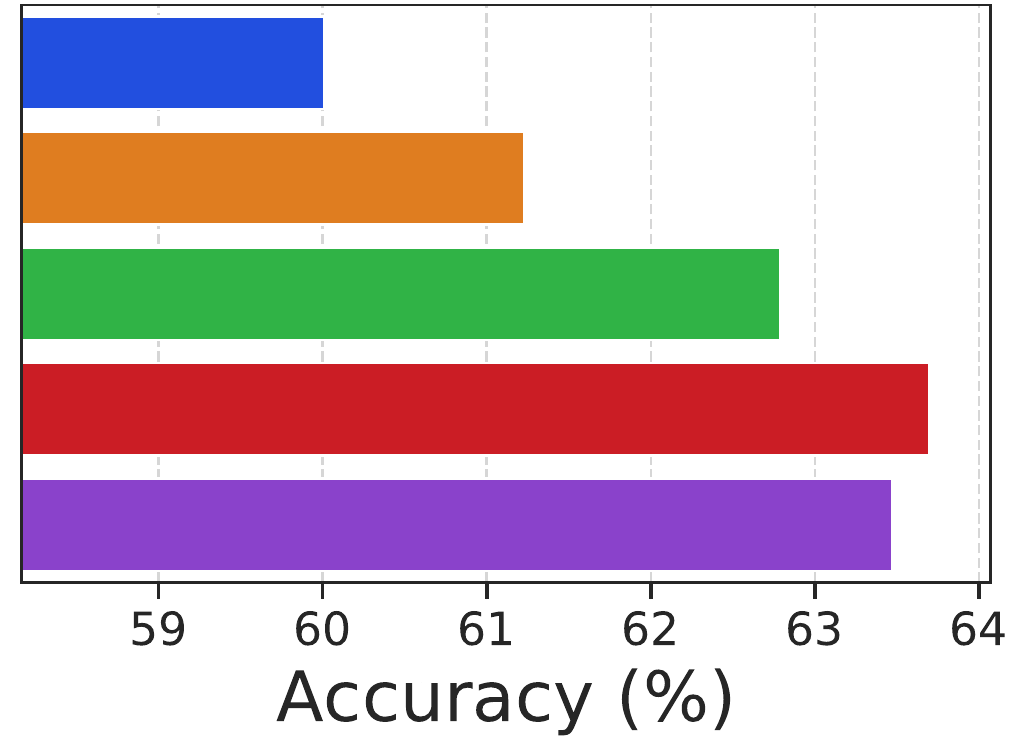}
        \caption{{\small Few-shot classification}}
        \label{fig:teaser2}
    \end{subfigure} \vspace{-2mm}
    \caption{\small \textbf{Average top-1 accuracy of different models on the downstream datasets.} (a) Linear evaluation with a fixed feature extractor and (b) 5-way 5-shot few-shot classification. In both cases, we observe that contrastive pretrained models achieve superior performance compared to cross-entropy pretrained models. Adding a self-supervised contrastive loss (\MOCO) improves the performance for both supervised cross-entropy and supervised contrastive pretrained models. We argue that incorporating a self-supervised contrastive loss (\MOCO) increases the variability within the same-class features and makes the models learn both high-level semantics and low-level cues.}
    \label{fig:teaser} 
    \vspace{-3mm}
\end{figure}

Self-supervised learning is an important research area whose goal is to learn superior data representations without any labelled supervision. Recently, self-supervised contrastive learning has shown promising results in image classification tasks 
\cite{he2020momentummoco, chen2020simplesimclr, caron2020unsupervisedswav}.
In the contrastive learning paradigm, a model is trained to recognize different augmentations of the same image (commonly referred as positives) while discriminating them from other random images (referred as negatives) in the dataset. The promising performance of self-supervised contrastive learning led to the idea of leveraging label information in the contrastive learning paradigm. To this end, Khosla \etal~\cite{khosla2020supervisedmit} proposed a supervised contrastive learning framework that achieves better ImageNet accuracy than the standard cross-entropy model.


Representations learned from contrastive learning have been shown to perform better than supervised cross-entropy models in various downstream tasks, particularly the object detection task \cite{he2020momentummoco, chen2020simplesimclr, khosla2020supervisedmit, tian2020makesgoodview, zhao2020makescuhk}. Despite recent progress, it is unclear why contrastive representations transfer better to other tasks, since most prior work focuses on in-domain evaluation, particularly ImageNet classification accuracy. In this paper, \textit{our goal is to understand the underlying mechanism of the superior transferability of contrastive learning}. Towards this end, we conduct a comprehensive study regarding transfer learning of contrastive approaches on downstream image classification, few-shot evaluation, and object detection. We rigorously benchmark five methods with different training objective losses: cross-entropy, self-supervised contrastive, supervised contrastive, joint cross-entropy/self-supervised contrastive, and joint supervised/self-supervised contrastive. 

We first compare the transfer performance of different ImageNet pretrained models on a collection of 12 downstream datasets from various domains. We find that contrastive methods perform much better than the supervised cross-entropy models, particularly in fixed feature transfer learning; however, the performance gap becomes smaller after full-network fine-tuning. We observe similar trends on other downstream tasks, including few-shot recognition and object detection and instance segmentation on the VOC0712 \cite{everingham2010pascal} and MS COCO \cite{lin2014coco} datasets. In particular, our results indicate that the joint objective of self-supervised contrastive loss and supervised cross-entropy/contrastive loss consistently outperforms the standard trained counterparts in different downstream tasks. Figure \ref{fig:teaser} shows the average top-1 accuracy of the different ImageNet pretrained methods we studied on the downstream datasets, for both fixed-feature linear evaluation and few-shot classification. Both the self-supervised contrastive model (denoted \MOCO) and supervised contrastive model (denoted \MIT) perform better than the cross-entropy model (denoted \CE). Moreover, the combination of cross-entropy and self-supervised contrastive (denoted \CEplusMOCO) performs better than cross-entropy or self-supervised contrastive alone. The same goes for the combination of self-supervised contrastive and supervised contrastive (denoted \MITplusMOCO).

We next investigate why contrastive approaches show superior transferability by analyzing the similarity between hidden representations, intra-class separation, and robustness to image corruption. We find that contrastive approaches learn more low-level and mid-level information that can be easily adapted to a different domain than the supervised cross-entropy models, which mostly learns high-level semantics in the penultimate layers. Zhao \etal~\cite{zhao2020makescuhk} hypothesized that one of the limiting factors of supervised cross-entropy models is the objective of minimizing intra-class variation. Our analysis also suggests that a model should have sufficient intra-class variation in the source domain to better transfer the learned representations to a different domain. Most standard supervised loss functions aim to increase inter-class distance and decrease intra-class variation, which might be harmful for transferability of features. We infer that contrastive approaches have larger within-class separation than the standard cross-entropy models, which could be one of the factors underlying their superior transferability. We also analyze the robustness and calibration of different models, and find that contrastive losses are more robust to different image corruptions and predict well-calibrated class probabilities that are more representative of true correctness likelihoods than cross-entropy models. Our key contributions in this work are as follows:
\begin{itemize}
\setlength\itemsep{-1pt}

    \item We benchmark five methods including cross-entropy, self-supervised contrastive, supervised contrastive, and their combinations on downstream image classification, object detection, and few-shot recognition. All results show a similar trend that contrastive learning extracts better features for transfer learning.
    
    \item We show that combining supervised loss with self-supervised contrastive loss improves transfer learning performance. Specifically, learned representation from the joint objective of self-supervised contrastive and supervised contrastive loss significantly outperforms the model trained with cross-entropy by 5.63\% under linear evaluation protocol and 3.46\% in few-shot recognition (5-shot) on the 12 downstream datasets, 1.37\% AP50 under object detection on VOC0712, and $\sim$0.8\% on MS COCO. The improvement of the joint objective over supervised contrastive model is small but consistent across all downstream tasks. The joint objective of cross-entropy and self-supervised contrastive loss also consistently performs better than the models trained with the individual objectives. 
    
    \item We apply Centered Kernel Alignment (CKA)~\cite{kornblith2019similarityCKA} and show that contrastive models contain more low-level and mid-level information in the penultimate layers than standard cross-entropy models. Furthermore, our analysis suggests that the contrastive models have higher intra-class variation than the standard cross-entropy models, even if the network is not explicitly trained to increase intra-class distance. 
    

\end{itemize}

\section{Related Work}
\label{sec:related_works}

\noindent
\textbf{Transfer Learning.} 
Early results on transfer learning showed that convolutional neural networks (CNNs) trained on large-scale datasets could be used to extract features to train SVMs and logistic regression models that outperformed hand-crafted feature-based approaches \cite{chatfield2014returnearly1,donahue2014decafearly2,sharif2014cnnearly3}. Transfer learning can be a powerful tool to train a significantly smaller dataset than the base dataset without overfitting. However, the factors driving the performance are still not completely understood. Huh \etal~\cite{huh2016makeshae16} investigated the effect of the source dataset on transfer learning. Simon \etal~\cite{kornblith2019betterimagenetmodel} found that pretrained models with higher ImageNet accuracy also tend to perform well in the downstream task. Azizpour \etal~\cite{azizpour2015factors} investigated the effect of network depth on the transfer performance. In this work, we show that contrastive training can improve transfer learning performance, and investigate the underlying principle behind it.

\vspace{1mm}
\noindent \textbf{Self-Supervised and Supervised Contrastive Learning.}
Earlier work on self-supervised learning generated pseudo labels by patch position \cite{Doersch_2015_ICCV}, image colorization \cite{zhang2016colorful}, image inpainting \cite{pathak2016context}, rotation \cite{gidaris2018unsupervised}, predictive coding \cite{henaff2019datacpc, henaff2019datacpc} and other pretext tasks.
Recently, contrastive learning has led to significant performance enhancement in self-supervised image representation learning. 
In particular, MoCo \cite{he2020momentummoco}, SimCLR \cite{chen2020simplesimclr}, SwAV \cite{caron2020unsupervisedswav}, and others have shown dramatic improvement in representation quality learned from unlabeled ImageNet images. Khosla \etal~\cite{khosla2020supervisedmit} proposed a new contrastive loss to leverage the label information. 
Moreover, representations learned from contrastive learning have been shown to perform better than supervised cross-entropy models in various downstream tasks \cite{zhai2019s4l,he2020momentummoco, chen2020simplesimclr, khosla2020supervisedmit,NEURIPS2020_ba7e36c4robustpretraining,NEURIPS2020_c39e1a03metaanalog}. However, most of the studies perform limited comparison, particularly with regard to fixed-feature transfer, few-shot learning and robustness, and the underlying principle of why contrastive learning transfers better still remains unclear.

\section{Analysis Setup}
\label{sec:analysis_setup}
Given a source domain $\mathcal{D}_s=\{(\mathbf{x}_{S_1}, y_{S_1}),(\mathbf{x}_{S_2}, y_{S_2}),\dots,(\mathbf{x}_{S_N}, y_{S_N})\}$ with a marginal distribution $\mathcal{P}_S$ and a target domain $\mathcal{D}_T=\{(\mathbf{x}_{T_1}, y_{T_1}),(\mathbf{x}_{T_2}, y_{T_2}),\dots,(\mathbf{x}_{T_N}, y_{T_N})\}$ with a  marginal distribution $\mathcal{P}_T$, where $(\mathbf{x}_{i}, y_{i})$ is the image-label pair, and, in general, $\mathcal{P}_S \ne \mathcal{P}_T$, the objective of transfer learning is to learn a target prediction function $f_T(\cdot)$ using the knowledge of $\mathcal{D}_S$. 
We study various target prediction tasks, namely, linear evaluation over fixed network for image classification, full-network fine-tuning for image classification, object detection, and few-shot image classification tasks.

\subsection{Loss Functions} \label{sec:loss}
\noindent \textbf{Supervised Cross-Entropy Loss.}
Supervised cross-entropy loss \cite{bridle1990probabilisticcrossentropy} is the standard loss function for multi-class classification. Given an input image $\mathbf{x}$ and one-hot encoded target label $\mathbf{y}$, denote the output representation of the encoder network as $\mathbf{v}=f_\theta(\mathbf{x})$. The class logits are calculated as $\mathbf{l}=\mathbf{W}\mathbf{v}+\mathbf{b}$, where $\mathbf{W} \in \mathbb{R}^{K\times D}$ contains the weights  and $\mathbf{b} \in \mathbb{R}^K$ is the bias of the final linear layer. The supervised cross-entropy loss is defined as:
\begin{equation}\label{eq:ce}
    \mathcal{L}_{\text{\CE}}(\mathbf{l}, \mathbf{y})=-\sum_{i=1}^{K} {y_i} \log\left(\frac{\exp(l_i)}{\sum_{j=1}^{K} \exp(l_j)}\right)
\end{equation}

\noindent \textbf{Self-Supervised and Supervised Contrastive Loss.}
In the contrastive learning paradigm, the network is trained by distinguishing between similar and dissimilar instances. We use Momentum Contrast (MoCo) \cite{he2020momentummoco}, particularly MoCov2 \cite{chen2020improvedmocov2}, for studying the efficacy of contrastive representations for transfer learning. The encoder $f_\theta(\cdot)$ of MoCo is a convolutional neural network (CNN), followed by a multi-layer perceptron (MLP) head to embed the encoded features in a contrastive subspace. 
MoCo has two base networks; one is actively trained to extract query features, and the other is the moving average of the query encoder to extract positive and negative features (commonly known as keys). Denote the query as $\mathbf{q}$ and the set of keys in the queue as $\{\mathbf{k}_1,\mathbf{k}_2,\dots,\mathbf{k}_M\}$. Assuming the key denoted by $\mathbf{k}_{+}$ matches with the query $\mathbf{q}$, the objective of MoCo is:

\begin{equation}\label{eq:moco}
    \mathcal{L}_{\text{\MOCO}}(\mathbf{\mathbf{q}})=-\log \frac{\exp(\mathbf{q} \cdot \mathbf{k}_{+} / \tau)}
    {\sum_{j=1}^{M} \exp(\mathbf{q} \cdot \mathbf{k}_j / \tau)},
\end{equation}
where $M$ is the queue size. In self-supervised contrastive loss, the positive key $\mathbf{k}_{+}$ is obtained from the augmented view of the same input image.

To leverage label information in contrastive learning, we follow the loss function from \cite{khosla2020supervisedmit}, where the positives are sampled from the contrastive features of the same classes as the query. The supervised contrastive loss is:

\begin{figure} [tbp]
    \centering
    \begin{subfigure}{0.48\linewidth}
        \centering
        \includegraphics[width=1\linewidth]{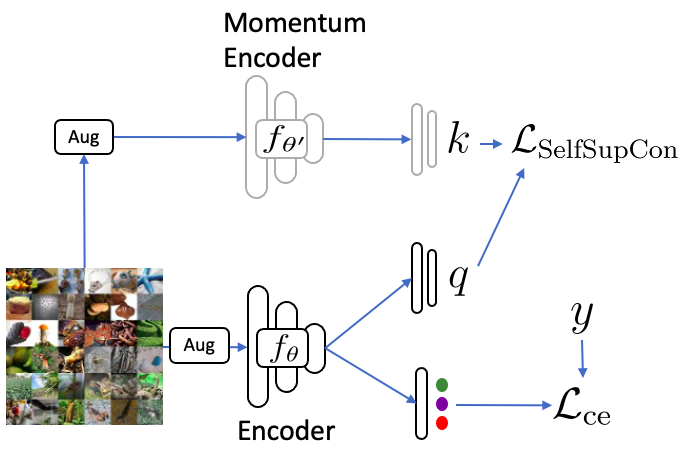}
        \caption{\CEplusMOCO}
        \label{fig:ce_plus_moco}
    \end{subfigure}
    \begin{subfigure}{0.48\linewidth}
        \centering
        \includegraphics[width=1\linewidth]{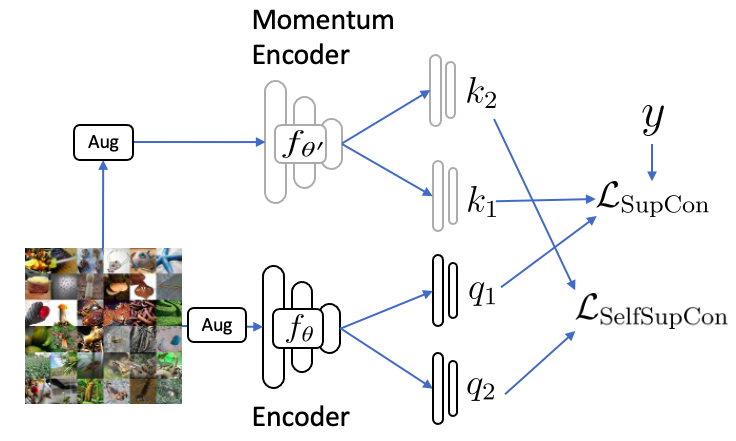}
        \caption{\MITplusMOCO}
        \label{fig:mit_plus_moco}
    \end{subfigure}
    \vspace{-2mm}
    \caption{\small Overview of the \CEplusMOCO and \MITplusMOCO models, both of which contain additional headers to optimize the corresponding loss terms.}
    \vspace{-3mm}
    \label{fig:method}
\end{figure}

\begin{equation}\label{eq:mit}
    \mathcal{L}_{\text{\MIT}}(\mathbf{\mathbf{q}})=
    \sum_{\substack{1 \le j \le M \\ y(\mathbf{q})=y(\mathbf{k_j})}}
    -\log \frac{\exp(\mathbf{q} \cdot \mathbf{k_j} / \tau)}
    {\sum_{j'=1}^{M} \exp(\mathbf{q} \cdot \mathbf{k_{j'}} / \tau)}.
\end{equation}


\subsection{Methods}
We examine five different methods to analyze the effectiveness of contrastive learning in tranferability, namely, \CE, \MOCO, \MIT, \CEplusMOCO, and \MITplusMOCO. For all the models, we use the same backbone (ResNet-50 for linear evaluation and fine-tuning; ResNet-18 for few-shot recognition) when evaluating to the downstream datasets. 
\CE, \MOCO, and \MIT are trained from the loss functions defined in Eq. \ref{eq:ce}, \ref{eq:moco}, and \ref{eq:mit}, respectively. \MOCO contains a ResNet backbone and an MLP projection head with two linear layers (2048-d hidden layer with ReLU and 128-d output layer). Note that the MLP projection head is only used during the pretraining stage. See \cite{chen2020improvedmocov2} for more details. \CEplusMOCO (Figure \ref{fig:ce_plus_moco}) contains two branches on top of the shared backbone; one branch minimizes the supervised cross-entropy loss ($\mathcal{L}_{\text{\CE}}$), and the other minimizes the self-supervised contrastive loss ($\mathcal{L}_{\text{\MOCO}}$). The supervised branch contains a linear layer to produce the class logits. The self-supervised branch consists of an MLP projection head containing two linear layers similar to MoCo. 
\MITplusMOCO (Figure \ref{fig:mit_plus_moco}) contains two headers with the shared backbone, one of which optimizes the self-supervised loss ($\mathcal{L}_{\text{\MOCO}}$), and the other minimizing the supervised contrastive loss ($\mathcal{L}_{\text{\MIT}}$). Both branches consist of an MLP projection header similar to MoCo. We also experimented with a single header instead of using two separate branches; however, it diverges during training.




\section{Experimental Results}
\label{sec:exp}

In the following sections, we describe the datasets, experimental setup, and results of our analysis. More details about datasets, implementation, and hyperparameter tuning are provided in Appendix \ref{ap:exp_details}.

\subsection{Datasets}
For source dataset training, we use ImageNet1K training set \cite{deng2009imagenet} with 1.28M images for downstream linear evaluation and full-network fine-tuning, and Mini-ImageNet~\cite{vinyals2016matchingnetworkminiimagenet} training set with 38k images for few-shot classification. For the downstream task of image classification, we use 12 datasets from different domains to evaluate the transferability of different models. The datasets are categorized as natural, satellite, symbolic, illustrative, medical, and texture.  Table \ref{tab:dataset} describes statistics of the benchmark datasets.
\begin{table}[t]
    \centering
    \begin{adjustbox}{max width=\linewidth}
    \begin{tabular}{l| l|r r r }
    \toprule
    Category & Dataset & Train Size & Test Size & Classes \\ 
    \midrule
    \multirow{3}{*}{Natural}
    & CropDisease~\cite{mohanty2016cropdisease} & 43456 & 10849 & 38 \\
    & Flowers~\cite{nilsback2008automatedflowers102} & 1020 & 6149 & 102 \\
    & DeepWeeds~\cite{DeepWeeds2019} & 12252 & 5257 & 9 \\
    \midrule
    \multirow{2}{*}{Satellite}
    & EuroSAT~\cite{helber2019eurosat} & 18900 & 8100 & 10 \\
    & Resisc45~\cite{cheng2017remoteresisc45} & 22005 & 9495 & 45 \\
    \midrule
    \multirow{2}{*}{Symbolic}
    & Omniglot~\cite{lake2015humanomniglot} & 9226 & 3954 & 1623 \\
    & SVHN~\cite{netzer2011readingsvhn} & 73257 & 26032 & 10 \\
    \midrule
    \multirow{2}{*}{Medical}
    & ISIC~\cite{codella2019skinisic} & 7007 & 3008 & 7  \\
    & ChestX~\cite{wang2017chestx} & 18090 & 7758 & 7 \\
    \midrule
    \multirow{2}{*}{Illustrative}
    & Kaokore~\cite{tian2020kaokore} & 6568 & 821 & 8 \\
    & Sketch~\cite{wang2019learningsketch} & 35000 & 15889 & 1000 \\
    \midrule
    Texture & 
    DTD~\cite{cimpoi2014DTD} & 3760 & 1880 & 47 \\
    \bottomrule
\end{tabular}
    \end{adjustbox}
    \vspace{-3mm}
    \caption{\small Datasets used for downstream image classification.}
    \vspace{-3mm}
    \label{tab:dataset}
\end{table}

\subsection{Experimental Setup}
We set the temperature parameter of MoCo to $\tau=0.07$, and queue size to 65596 for ImageNet pretraining and 16384 for Mini-ImageNet pretraining for all contrastive models. 
For both the cross-entropy and contrastive models, we use the standard data augmentations used in the literature for the respective models. For \CEplusMOCO, we use the same MoCo augmentation for both the cross-entropy and contrastive branches. We also provide ablations on \CE with MoCo augmentation in Sec.~\ref{sec:ablation_studies}. Unless otherwise mentioned, we use top-1 accuracy as the evaluation metric. When performing transfer learning to the downstream datasets, we created a separate validation set from the training set, and swept the hyperparameters (learning rate, batch size, and weight decay) for each dataset. Then we use the optimal hyperparameters to train the full training set (train+val), and evaluate on the test set.  We also perform multiple runs with different random seeds, and report mean score among different runs.  See Appendix \ref{ap:more_results_ab} for detailed results with confidence intervals over multiple runs.




\begin{figure}[tbp]
    \centering
    \includegraphics[width=1.\linewidth]{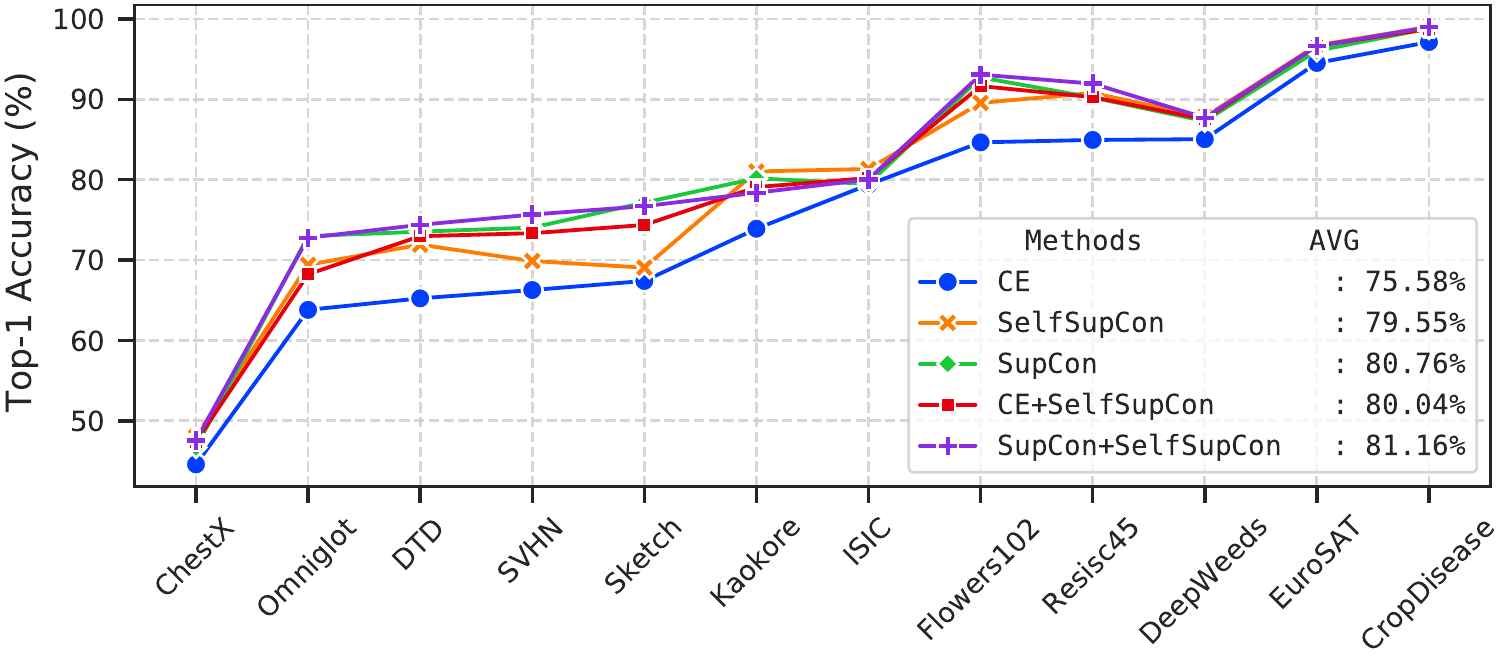} 
    \vspace{-7mm}
    \caption{\small Top-1 accuracy of different models on the downstream datasets for fixed-feature extractor transfer learning (averaged over 5 runs). The models are pretrained on the ImageNet1K dataset and we only train the final linear layer on top of the pretrained backbones. Best viewed in color.}
    \label{fig:fixed_linear} 
    \vspace{-2mm}
\end{figure}

\subsection{Linear Evaluation over Fixed Network}

We first use the linear evaluation over fixed network to test the learned visual representations. For all methods, we freeze the backbone, only train a linear layer on top of the backbone, and optimize the cross-entropy loss with the SGD optimizer. The learning rate, weight decay, and batch size are selected by hyperparameter tuning on the validation set. Figure~\ref{fig:fixed_linear} shows the performance of different models on the 12 downstream datasets in terms of top-1 accuracy (averaged over 5 runs). Note that our reproduced cross-entropy trained model performs slightly differently than the pretrained ResNet-50 model from PyTorch \cite{NEURIPS2019_9015pytorch}.

\begin{table*}[!tbh]
    \centering
    \begin{adjustbox}{max width=\linewidth}
    \begin{tabular}{l|llllllllllll|l}
\toprule
{} &             \rot{CropDisease} &               \rot{DeepWeeds} &               \rot{Flowers102} &                 \rot{EuroSAT} &                \rot{Resisc45} &                    \rot{ISIC} &                  \rot{ChestX} &                \rot{Omniglot} &                    \rot{SVHN} &                 \rot{Kaokore} &                  \rot{Sketch} &                     \rot{DTD} &                    \rot{Mean} \\
\midrule
\multicolumn{14}{l}{Full Dataset} \\
\midrule
\CE          &  99.92 &  97.37 &   93.47 &  98.78 &  96.03 &  88.06 &  55.67 &  90.34 &  97.03 &  88.13 &  79.49 &  73.26 &  88.13 \\
\MOCO        &  99.91 &  97.39 &  93.36 &  98.85 &  96.28 &  88.13 &  56.41 &  91.10 &  97.25 &  88.92 &  77.16 &  75.49 &  88.35 \\
\MIT         &  99.91 &  96.89 &   95.37 &  98.67 &  96.01 &  87.92 &  55.08 &  90.56 &  97.11 &  87.88 &  80.14 &  74.19 &  88.31 \\
\CEplusMOCO  &  99.88 &  97.28 &   95.30 &  98.91 &  96.18 &  88.29 &  54.77 &  90.20 &  97.14 &  88.25 &  80.85 &  74.12 &  {\bf88.43} \\
\MITplusMOCO &  99.91 &  97.38 &   96.17 &  98.75 &  96.21 &  88.55 &  55.12 &  90.86 &  97.04 &  87.73 &  80.17 &  74.86 &  {\bf88.56} \\
\midrule
\multicolumn{14}{l}{{1000 training samples}} \\
\bottomrule
\end{tabular}
    \end{adjustbox}
    \vspace{-2mm}
    \caption{\small  Performance of different models on the downstream datasets in terms of top-1 accuracy (\%) (averaged over 5 runs) for full-network fine-tuning. Contrastive pretrained methods are slightly more effective in a limited data regime than cross-entropy based models.}
    \label{tab:tfmbn_score_all}
\end{table*}

\begin{table*}[!tbh]
    \centering
    \begin{adjustbox}{max width=\linewidth}
        \begin{tabular}{l| ccc || c c c | c c c || c c c | c c c}
    \toprule
        Datasets & \multicolumn{3}{c||}{VOC0712} & \multicolumn{12}{c}{MS COCO (Trained with 1$\times$ schedule)} \\ 
        Detectors & \multicolumn{3}{c||}{FasterRCNN-R50-C4}  & \multicolumn{6}{c||}{MaskRCNN-R50-C4}  & \multicolumn{6}{c}{MaskRCNN-R50-FPN} \\ 
        Methods &  AP$^{\text{bb}}_\text{50}$ &  AP$^{\text{bb}}$ & AP$^{\text{bb}}_{\text{75}}$ &  AP$^{\text{bb}}$ &  AP$^{\text{bb}}_{\text{50}}$ & AP$^{\text{bb}}_{\text{75}}$ &  AP$^{\text{mk}}$ &  AP$^{\text{mk}}_{\text{50}}$ & AP$^{\text{mk}}_{\text{75}}$ &  AP$^{\text{bb}}$ &  AP$^{\text{bb}}_{\text{50}}$ & AP$^{\text{bb}}_{\text{75}}$ &  AP$^{\text{mk}}$ &  AP$^{\text{mk}}_{\text{50}}$ & AP$^{\text{mk}}_{\text{75}}$\\
    \midrule
    
        \CE & 81.58	&	54.63	&	60.17	&	38.91	&	59.05	&	42.14	&	33.98	&	55.39	&	36.10	&	39.84	&	60.54	&	43.60	&	36.54	&	57.65	&	39.25 \\ 
\MOCO & 82.48	&	57.33	&	64.03	&	39.14	&	58.74	&	42.40	&	34.29	&	55.53	&	36.62	&	39.19	&	59.54	&	42.74	&	36.03	&	56.79	&	38.75 \\ 
\MIT & 82.64	&	56.12	&	62.45	&	39.63	&	59.76	&	42.77	&	34.52	&	56.26	&	36.54	&	40.40	&	61.20	&	44.35	&	36.97	&	58.31	&	39.80 \\ 
\CEplusMOCO & \bf83.06	&	57.07	&	63.97	&	39.68	&	59.84	&	42.80	&	34.62	&	56.31	&	36.74	&	\bf40.65	&	61.38	&	44.65	&	\bf37.21	&	58.57	&	40.06 \\ 
\MITplusMOCO & 82.95	&	57.26	&	64.12	&	\bf39.95	&	60.02	&	43.12	&	\bf34.93	&	56.59	&	37.27	&	40.33	&	60.96	&	44.20	&	36.95	&	58.19	&	39.73 \\ 

    \bottomrule
    \multicolumn{16}{l}{\footnotesize{The results of VOC0712 is the average of 5 runs. AP$^{\text{bb}}$: AP of objection detection; AP$^{\text{mk}}$: AP of instance segmentation.}} \\
    \end{tabular}
    
    \end{adjustbox}
    \vspace{-3mm}
    \caption{\small Object detection and instance segmentation results on VOC0712 and MS COCO (averaged over 5 runs).}
    \label{tab:od_1x}
    \vspace{-2mm}
\end{table*}

We note that the cross-entropy model performs the worst as a fixed feature extractor, both in terms of the accuracy for an individual dataset and the final average across all datasets. All of the contrastive approaches perform better than cross-entropy. We also find that training the cross-entropy model with strong augmentation as MoCo does not help much either. The best performing model \MITplusMOCO performs, on average, { $\textbf{5.63}\%$} better than \CE. For datasets that are different than ImageNet, e.g., SVHN, Sketch, Omniglot, and DTD, \CE performs much worse than contrastive models. We infer that features learned from the cross-entropy model are not directly helpful for datasets that are much different than ImageNet, while features learned from contrastive approaches are applicable to the datasets from different domains. Moreover,  \CEplusMOCO performs better than \CE or \MOCO and \MITplusMOCO performs better than the individual \MIT or \MOCO model, suggesting that \emph{self-supervised contrastive learning improves the transferability of supervised cross-entropy and supervised contrastive learning}.

\subsection{Full-Network Fine-Tuning}
We further fine-tune the full network to study the transferability of all the methods. Here we include the results of image classification and object detection.

\vspace{1mm}
\noindent\textbf{Image Classification.}
We fine-tune the pretrained models along with the final linear header on  downstream datasets. We perform different hyperparameter sweeping on the validation set and report the scores on the test set for the optimal hyperparameters. Table~\ref{tab:tfmbn_score_all} shows the top-1 accuracy of all datasets for full-network fine-tuning for 5 runs. While \CE performs much worse than other contrastive models in the linear evaluation experiments, we did not observe similar behavior for full-network fine-tuning. \MITplusMOCO achieves $88.56\%$ top-1 accuracy, just $0.43\%$ better than \CE, which achieves $88.13\%$ top-1 accuracy. We also report the performance of fine-tuning with only 1000 training samples, where \CEplusMOCO performs $1.09\%$ better than \CE, and \MITplusMOCO performs $1.48\%$ better than \CE. We infer that contrastive pretrained methods are slightly more effective in a limited data regime than cross-entropy based models. However, when we have a sufficient amount of data, all models achieve similar performance.  

\vspace{1mm}
\noindent \textbf{Object Detection and Instance Segmentation.} 
We conduct experiments on object detection and instance segmentation on VOC0712 \cite{everingham2010pascal} and MS COCO \cite{lin2014coco} to validate the learned representations from different models based on Detectron2 \cite{wu2019detectron2}. We follow the settings in \cite{he2020momentummoco} to fine-tune the whole network while only training a few epochs (1$\times$ schedule in the Detectron2 setting).
The results are shown in Table \ref{tab:od_1x}. \MIT provides slightly better results than \CE and \MOCO. Furthermore, \CEplusMOCO and \MITplusMOCO achieved consistent improvement on AP over their counterparts by $\sim$0.8\% and $\sim$0.5\% on MS COCO, respectively. Again, the results echo our previous observations in linear evaluation experiments.


\begin{table*}[!th]
    \centering
    \begin{adjustbox}{max width=\linewidth}
    \begin{tabular}{l|l||llllllllllll|l}
\toprule
{} & \rot{Mini-IN$^*$} & \rot{CropDisease} & \rot{DeepWeeds} & \rot{Flowers102} & \rot{EuroSAT} & \rot{Resisc45} &   \rot{ISIC} & \rot{ChestX} & \rot{Omniglot} &   \rot{SVHN} & \rot{Kaokore} & \rot{Sketch} &    \rot{DTD} &   \rot{Mean} \\
\midrule
\multicolumn{15}{l}{{5-shot}} \\
\midrule
\CE          &  72.47 &  86.58 &  48.33 &  81.31 &  78.51 &  72.86 &  44.28 &  26.16 &  94.26 &  27.98 &  37.63 &  64.64 &  57.64 &  60.01 \\
\MOCO        &  67.71 &  83.29 &  49.90 &  84.29 &  81.65 &  72.36 &  45.20 &  26.91 &  93.61 &  27.10 &  42.75 &  66.32 &  61.35 &  61.23 \\
\MIT         &  75.20 &  83.44 &  47.74 &  82.93 &  80.94 &  74.48 &  42.97 &  26.23 &  96.78 &  33.22 &  45.10 &  74.24 &  65.42 &  62.79 \\
\CEplusMOCO  &  76.13 &  84.68 &  50.02 &  86.88 &  82.63 &  75.11 &  44.66 &  27.93 &  96.19 &  31.36 &  45.32 &  72.38 &  67.21 &  {\bf63.70} \\
\end{tabular}
    \end{adjustbox}
    \vspace{-2mm}
    \caption{\small Few-shot classification accuracies (average score over 600 episodes) for 5-shot and 20-shot on the Mini-ImageNet and 12 downstream datasets. Contrastive approaches consistently perform better than cross-entropy across all downstream datasets.}
    \label{tab:fewshot_1520}
\end{table*}

\begin{figure*}[!tbh]
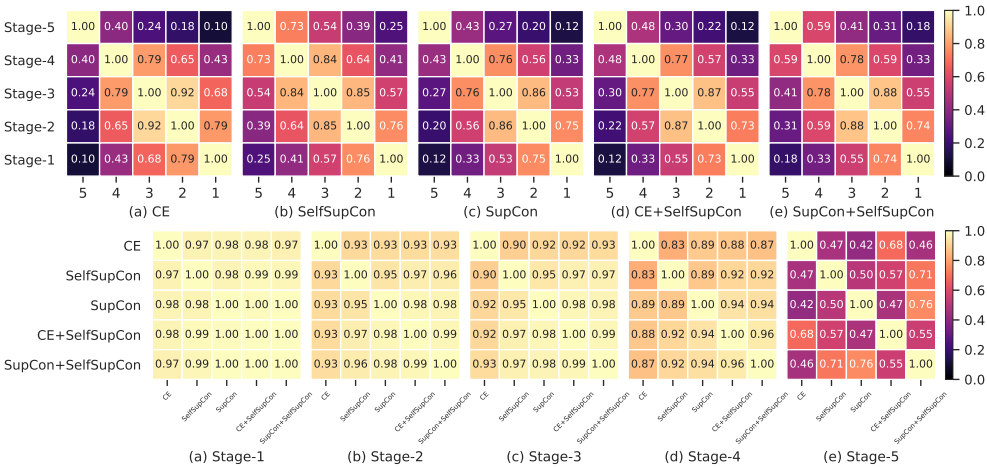

  \begin{minipage}[c]{0.75\textwidth}
    \adjustimage{width=1\columnwidth,right}{Figures/cka_intra.pdf} \\
    \adjustimage{width=1\columnwidth,right}{Figures/cka_inter.pdf}
  \end{minipage}\hfill
  \begin{minipage}[c]{0.22\textwidth}
    \caption{
      \small Centered Kernel Alignment (CKA) scores. Top: between different stages of the same model, which shows that final layers of a contrastive model are more similar to initial layers than in the cross-entropy model. Bottom: the same stage between different models, showing that different models learn similar representation in the initial layers but diverge drastically in the final layers.
    } \label{fig:CKA}
  \end{minipage}
  \vspace{-2mm}
\end{figure*}

\subsection{Few-shot Classification}
For few-shot learning, we use the ResNet-18 backbone, and pretrain all models on the Mini-ImageNet dataset. As suggested by \cite{tian2020rethinkingfewshotgoodembedding}, a model providing good embedding is essential for few-shot learning; thus, we simply trained a logistic regression classifier on top of the fixed network for the few-shot classification. Table \ref{tab:fewshot_1520} shows the average top-1 accuracy of 600 episodes for 5-way 5-shot and 20-shot experiments.
We observe a similar trend that contrastive approaches consistently perform better than cross-entropy across all downstream datasets. \CEplusMOCO and \MITplusMOCO achieve the best scores, which suggests that self-supervised contrastive learning improves upon both supervised cross-entropy and supervised contrastive learning. On the other hand, when performing in-domain few-shot on the Mini-ImageNet test set, we observe that \MITplusMOCO performs worse than \MIT, suggesting that \emph{the performance of same-domain few-shot classification might not be a good proxy for cross-domain few-shot performance}.

\section{Discussion and Analysis}
\label{sec:ablation}

\noindent
\textbf{Contrastive approaches learn more low/mid-level features.} 
Figure \ref{fig:CKA} (top row) shows the similarity between different stages of the same ResNet-50 model in terms of the centered kernel alignment (CKA) \cite{kornblith2019similarityCKA}. The initial stages mostly learn low-level features, while the final stages learn more semantic information. The \CE model has the least similarity between the representations of the final ResNet stage and the initial stage with a CKA score of 0.10. We infer that the final ResNet stages of \CE contain mostly domain-specific high-level semantics.  While \MIT is also trained in a supervised fashion to increase inter-class distance, it has a higher CKA score of 0.12 between the final and initial layers, which suggests that the final layers of \MIT contain more low-level and mid-level information than \CE. As expected, the final stage of \MOCO contains the most low/mid-level information, and \CEplusMOCO and \MITplusMOCO show higher CKA scores between the final layers and initial layers than their supervised counterparts. 

Similarly, Figure~\ref{fig:CKA} (bottom row) shows the similarity between representations of the same ResNet stages between different models. We observe that the representations from ResNet stage-1 to stage-3 are highly similar between different methods, with CKA scores more than 0.9. However, the representations become slightly differentiated in ResNet stage-4 and highly dissimilar in the final ResNet stage. For example, the CKA between \CE and \MOCO in the final layer is only 0.47. Surprisingly, CKA between \CE and \MIT is 0.42 despite the fact that both models were trained with ImageNet1k with full supervision, which suggests that \MIT learns a much different feature representation than \CE. Moreover, all contrastively trained models are more similar to each other than to cross-entropy model.

\begin{figure}[!tb]
    \centering
    \begin{subfigure}{0.30\linewidth}
    \includegraphics[width=2.5cm]{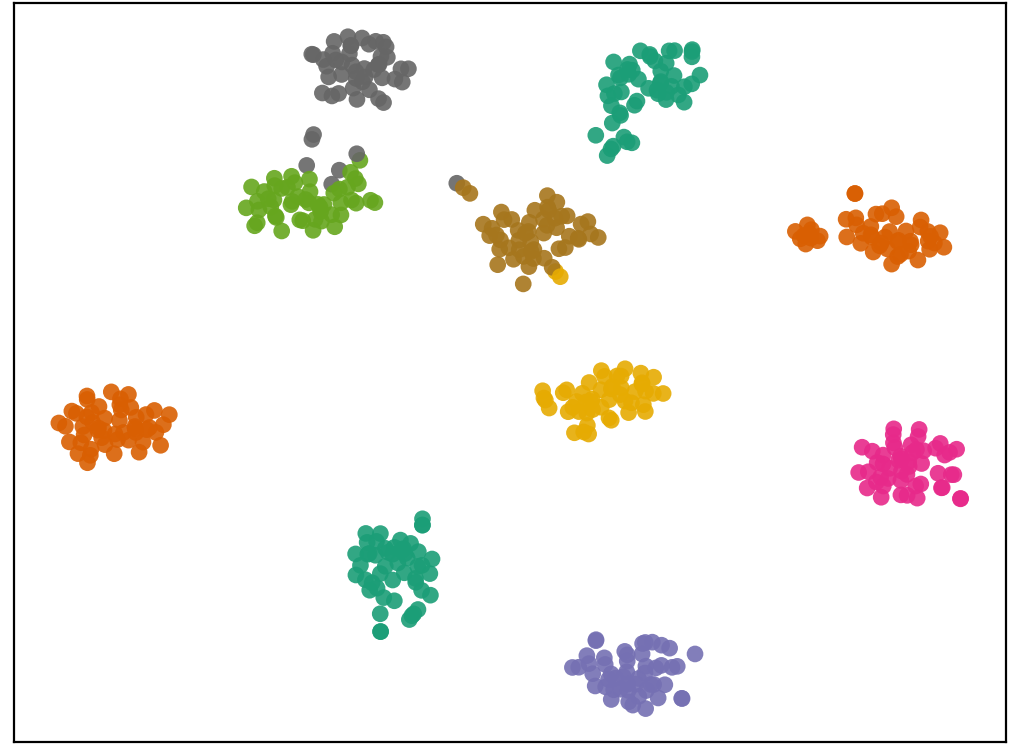}
    \caption{\scriptsize CE}
    \end{subfigure}
    \begin{subfigure}{0.30\linewidth}
    \includegraphics[width=2.5cm]{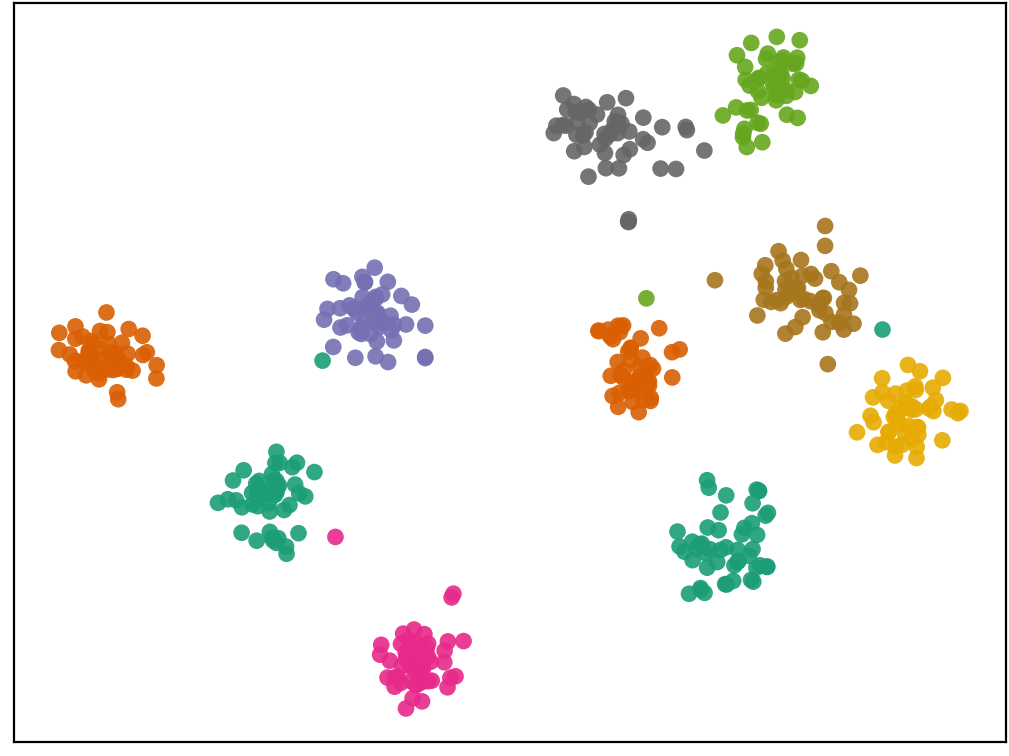}
    \caption{\scriptsize CE+SelfSupCon}
    \end{subfigure}
    \begin{subfigure}{0.30\linewidth}
    \includegraphics[width=2.5cm]{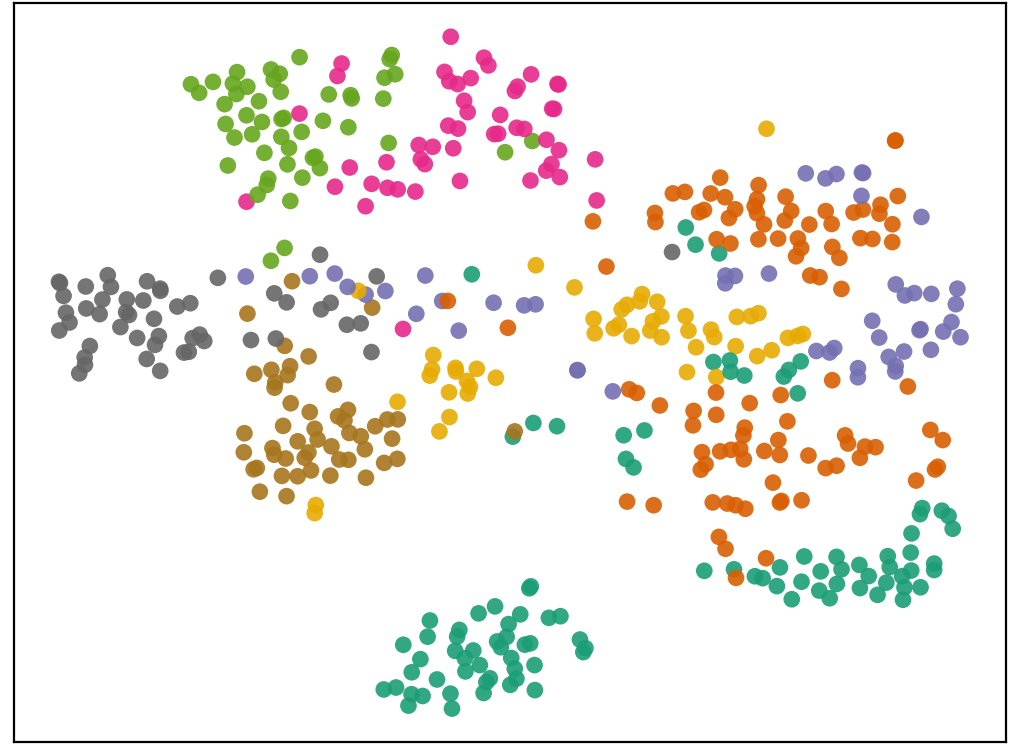}
    \caption{\scriptsize SelfSupCon}
    \end{subfigure}\\ 
    \begin{subfigure}{0.30\linewidth}
    \includegraphics[width=2.5cm]{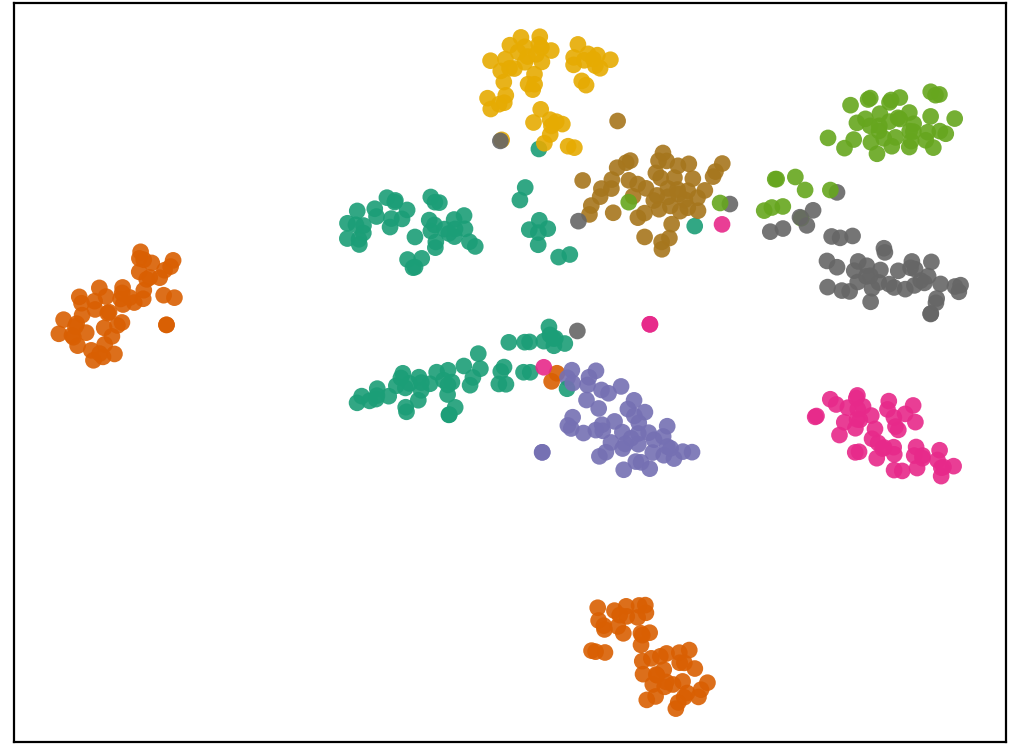}
    \caption{\scriptsize SupCon}
    \end{subfigure}
    \begin{subfigure}{0.35\linewidth}
    \includegraphics[width=2.5cm]{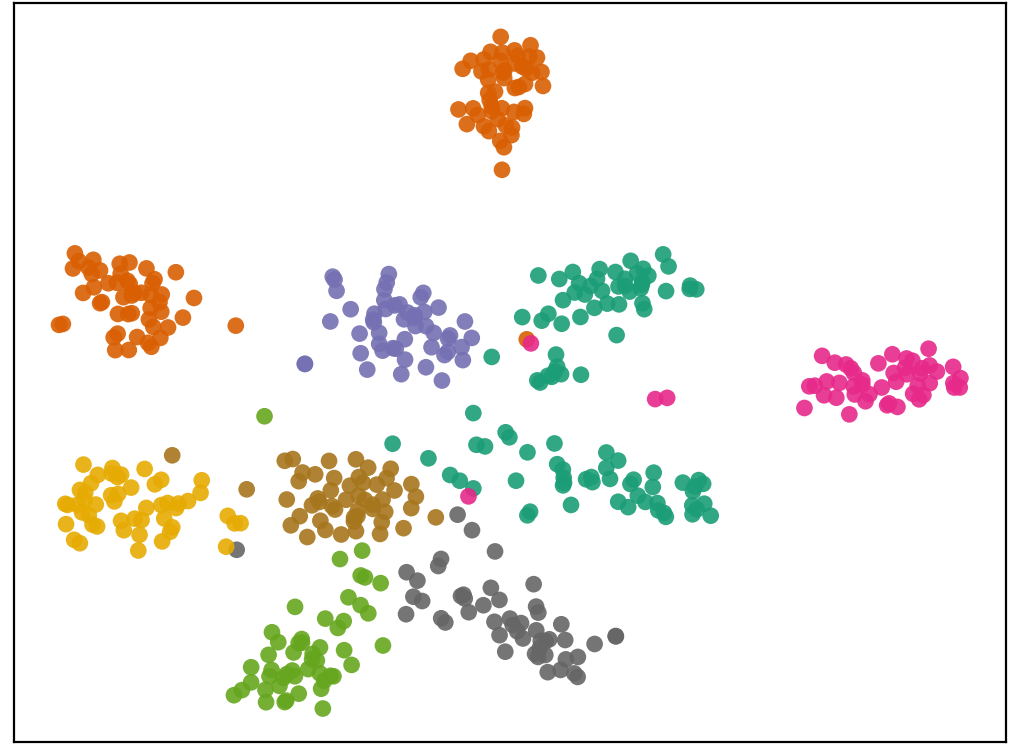}
    \caption{\scriptsize SupCon+SelfSupCon}
    \end{subfigure}
    \vspace{-2mm}
    \caption{\small t-SNE visualization of penultimate layer activations of different models on the ImageNet dataset. The projections of \CE are much tighter, whereas the projections of contrastive approaches are spread into broad clusters. Best viewed in color.}
    \label{fig:tsne} 
    \vspace{-4mm}
\end{figure}

Supervised learning models learn feature representations using objectives that also increase the inter-class separation. However, we argue that increasing the intra-class variation, though possibly harmful for in-domain performance, is beneficial for learning rich feature representations in transfer learning. t-SNE visualizations in Figure \ref{fig:tsne} show that the clusters in contrastive methods are more spread out than in vanilla cross-entropy, which also supports our claim.

\vspace{1mm}
\noindent \textbf{Contrastive models improve model calibration.}
There are several metrics for measuring the models' calibration. Here we adopt the Expected Calibration Error (ECE) and Negative Log Likelihood (NLL)~\cite{guo2017calibration}. See Appendix \ref{ap:in_derivative} for the experimental setup.
Table \ref{tab:res_on_IN_val} reports the performance for ImageNet pretrained models on the ImageNet validation dataset in terms of Top-1 accuracy (higher is better), Negative Log Likelihood (lower is better), and Expected Calibration Error (lower is better). We see that well-calibrated models do not necessarily have higher accuracy.  In particular, \MOCO shows the best calibration performance, but worst top-1 accuracy score. Moreover, \CEplusMOCO has better ECE and NLL scores than either \CE or \MOCO, and \MITplusMOCO shows better calibration scores than \MIT or \MOCO. We also evaluate the ImageNet pretrained models on the Stylized ImageNet validation set \cite{geirhos2018SIN} to see whether contrastive approaches learn both texture-based and shape-based representations. In Table \ref{tab:res_on_IN_val}, we show that contrastively trained models perform better on the Stylized ImageNet validation set than the cross-entropy model. While \MITplusMOCO has slightly lower accuracy than \MIT, it improves the calibration by $2.93\%$. Note that \MOCO performs the worst in terms of top-1 accuracy, which is expected since the backbone has not been trained with label information, and both of the datasets in Table~\ref{tab:res_on_IN_val} contain ImageNet classes. However, \MOCO performs better than \CE in terms of ECE score. Overall, our experiments suggest that contrastive approaches produce more calibrated predictions than the cross-entropy model on both in-domain evaluation and transfer learning. Calibration performance on the 12 downstream datasets is provided in Appendix \ref{ap:calib_downstream}. \\


\begin{table}[tb]
    \centering
    \begin{adjustbox}{max width=\linewidth}
    \begin{tabular}{l|c c c| c c}
\toprule
           & \multicolumn{3}{c|}{ImageNet} & \multicolumn{2}{c}{Stylized ImageNet} \\
    Method & Top-1(\%) & NLL & ECE(\%) & Top-1(\%) & ECE(\%) \\
\midrule
    \CE &  76.60 & 0.9575 & 6.63 & 7.50 & 33.19 \\
    \MOCO & 66.07 & 1.3943 & {\bf1.60} & 5.34 & 21.59 \\
    \MIT & {\bf77.32} & 0.9466 & 4.49 & {\bf10.12} & {\bf16.35} \\
    \CEplusMOCO & {\bf77.47} & {\bf0.9106} & 5.14 & {\bf9.77} & 24.36  \\
    \MITplusMOCO & 77.11 & {\bf0.9306} & {\bf4.31} & 9.70 & {\bf13.42}  \\
\bottomrule
\end{tabular}
    \end{adjustbox}
    \caption{\small Performance on ImageNet validation in terms of Top-1 accuracy (higher is better), Negative Log Likelihood (lower is better), and Expected Calibration Error (lower is better). All the models were trained on the ImageNet1K training dataset.}
    \label{tab:res_on_IN_val}
\end{table}

\vspace{5mm}
\noindent
\textbf{Contrastive learning is robust to image corruption.}
Many deep learning models lack robustness to natural corruptions. 
In Table~\ref{tab:robust_main}, we report the robustness performance of different models on the ImageNet-R \cite{hendrycks2020manyimagenetr}, ImageNet-A  \cite{hendrycks2019naturaladversarialimageneta}, and ImageNet-C \cite{hendrycks2019benchmarkingimagenetc} datasets. 
Contrastive approaches, with the exception of \MOCO, show superior performance for both the ImageNet-A and ImageNet-R datasets, in both top-1 accuracy and expected calibration error (ECE) \cite{guo2017calibration}. The improvement is particularly noticeable between \CE and \CEplusMOCO. \CEplusMOCO improves the accuracy over \CE by 5.18\% for ImageNet-A and 4.38\% for ImageNet-R, and lowers the calibration error by 1.58\% for ImageNet-A and 3.72\% for ImageNet-R. 
\begin{table}[tb!]
    \centering
    \begin{adjustbox}{max width=\linewidth}
    \begin{tabular}{l|c c | c c| c}
\toprule
    \multirow{2}{*}{Method} & \multicolumn{2}{c|}{ImageNet-R} & \multicolumn{2}{c|}{ImageNet-A} & ImageNet-C \\ 
    & Top-1(\%) & ECE(\%) & Top-1(\%) & ECE(\%) & mCE(\%) \\
\midrule
    \CE &  35.83 & 19.45 & 3.35 & 55.03 & 60.80 \\
    \MOCO & 26.51 & 25.14 & 2.85 & 48.47 & 68.87  \\
    \MIT & {\bf42.01} & {\bf13.40} & {\bf7.36} & {\bf46.78} & {\bf54.23} \\
    \CEplusMOCO & {\bf41.01} & 17.87 & {\bf7.71} &  51.19 & {\bf54.27} \\
    \MITplusMOCO & 38.57 & {\bf13.60} & 6.36 & {\bf46.16} & 54.40 \\
\bottomrule
\end{tabular}
    \end{adjustbox}
    \caption{\small Robustness tests on the ImageNet-R, ImageNet-A, and ImageNet-C datasets. ECE is the expected calibration error (lower is better) and mCE (lower is better) is the mean of the (unnormalized) corruption errors of the Noise, Blur, Weather, and Digital corruptions. Models are trained only on clean ImageNet images.}
    \label{tab:robust_main}
    \vspace{-2mm}
\end{table}
The rightmost column of Table \ref{tab:robust_main} reports performance of different models on ImageNet-C in terms of (unnormalized) mean corruption error (mCE) of the Noise, Blur, Weather, and Digital corruptions. Lower mCE denotes that the model is more robust to different corruption types. All the models are trained on clean ImageNet1K dataset. We observe that \CEplusMOCO, \MIT, and \MITplusMOCO perform the best across different models, and also provide better representations that are transferable to different domains, as shown in the linear evaluation and few-shot experiments. We also note that there is no single contrastive model that works best for all metrics in terms of robustness; however, contrastive loss, in general, improves the neural network robustness.  \\

\section{Ablation Studies} \label{sec:ablation_studies}

\begin{figure}[!tb]
\vspace{-2mm}
    \centering
    \includegraphics[width=0.4\textwidth]{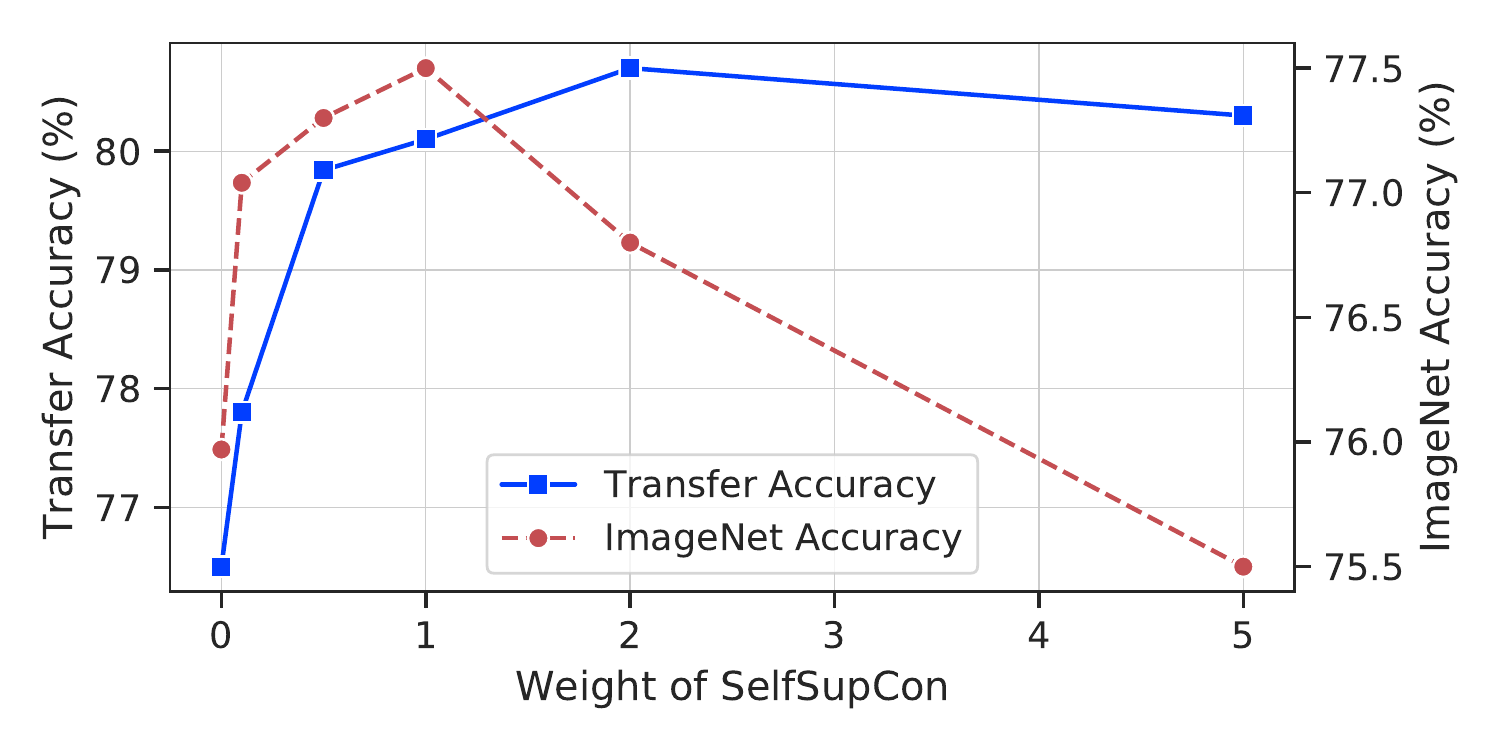}
    \vspace{-0.3cm}
    \caption{\small Effect of different weights on the $\mathcal{L}_{\text{\MOCO}}$ term in the \CEplusMOCO model. Best viewed in color.}
    \label{fig:weight_moco}
\end{figure}

\noindent
\textbf{Ablations on weights of \MOCO loss for model with joint objective.}
As described in Section~\ref{sec:loss}, \CEplusMOCO is trained with the objective of $\mathcal{L}_{\text{\CE}} + \alpha \mathcal{L}_{\text{\MOCO}}$, where $\alpha$ is the weight on self-supervised contrastive loss. Figure~\ref{fig:weight_moco} reports the effect of $\alpha$ on average transfer accuracy of 12 downstream datasets and ImageNet1K validation accuracy. We obtain the highest ImageNet accuracy for $\alpha=1$; however, the highest transfer accuracy is reported at $\alpha=2$, which indicates that higher ImageNet accuracy does not always imply higher transfer accuracy. Higher values of $\alpha$ impose more intra-class variation, and there is an optimal value where transferability of the model is maximized. Imposing more intra-class distance might hurt  transfer performance, as shown in figure that when $\alpha > 2$ the transfer accuracy gradually decreases. 

\vspace{1mm}
\noindent \textbf{Effect of augmentations on \CE model.}
For the \CE model, we adopt standard data augmentation as used in ResNet-50 ImageNet training. We also train a cross-entropy model with additional augmentations, such as color-jitter, random gray-scale, and Gaussian blur, so that all the models are pre-trained with similar augmentations. We denote this model as \CEstrong. Table~\ref{tab:ce_strong_mean} shows mean accuracy of 12 downstream datasets for \CEstrong in linear evaluation, full-network fine-tune, and few-shot classification (averaged over 5 runs). \CEstrong is just slightly better than \CE for transfer learning; however, contrastive approaches are still significantly better than both \CE and \CEstrong, particularly for fixed-feature transfer.

\begin{table}[tb]
\fontsize{7.5}{9}\selectfont
    \centering
    \begin{tabular}{c|c c c}
    \toprule
    Method & Linear-Evaluation & Finetune & Few-shot \\  
    \midrule
    \CE & 75.67 & 88.13 & 60.01 \\
    \CEstrong & 75.91 & 88.27 & 61.31  \\
    \bottomrule
    \end{tabular}
    \caption{\small Performance comparison between \CE and \CEstrong for linear evaluation, full-network fine-tuning, and 5-shot few-shot classification in terms of average top-1 accuracy over the 12 downstream datasets.}
    \label{tab:ce_strong_mean}
    \vspace{-2mm}
\end{table}

\begin{figure}[tb]
    \centering
    \includegraphics[width=0.48\textwidth]{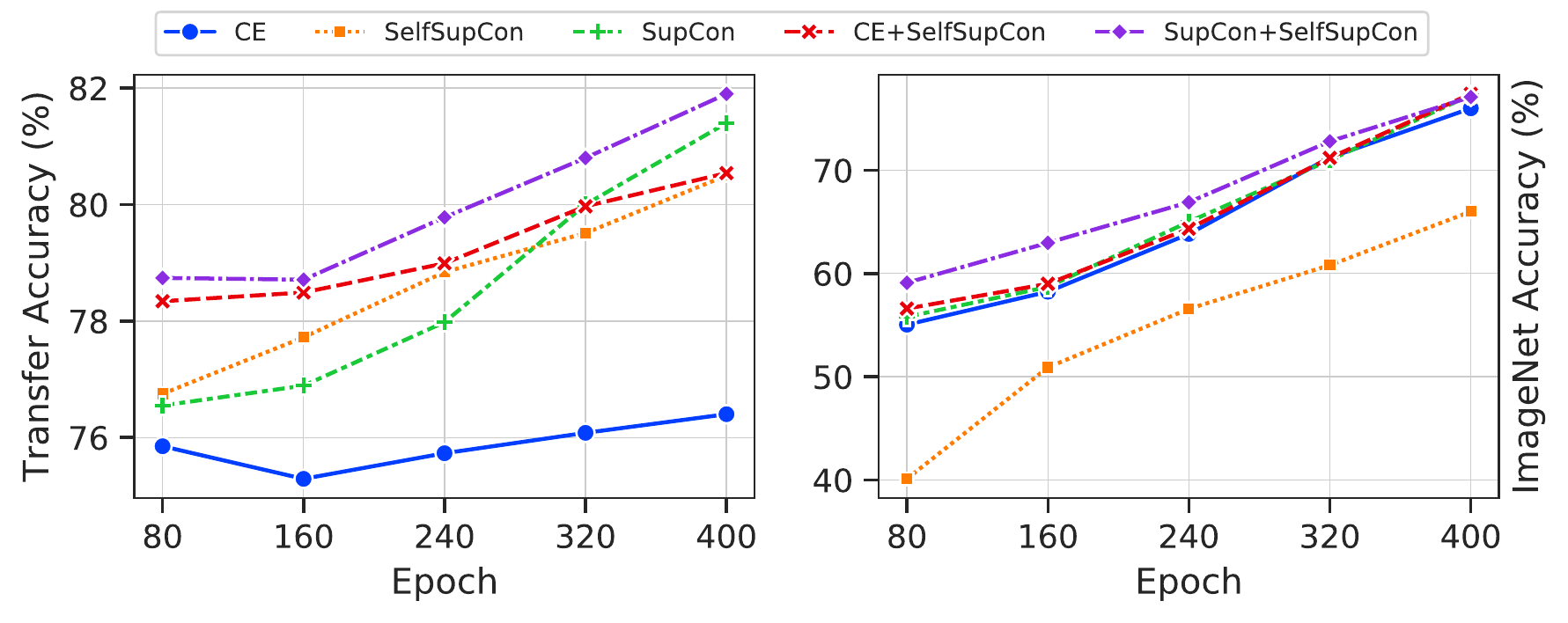}
    \vspace{-5mm}
    \caption{\small Left: Average linear evaluation accuracy (\%) of all the downstream datasets, and Right: ImageNet validation accuracy, for  intermediate checkpoints from different ImageNet-pretraining epochs. Transferable representations from the cross-entropy model do not improve much with more pretraining epochs on ImageNet. However, for contrastive approaches, longer pretraining improves transferability. Best viewed in color.}
    \label{fig:epoch_vs_transfer}
    \vspace{-3mm}
\end{figure}

\vspace{1mm}
\noindent
\textbf{Does transferability improve with longer training?}
We study the effect of pretrained checkpoints from different ImageNet-training epochs on the transferability of visual representations. Figure \ref{fig:epoch_vs_transfer} shows average transfer accuracy for fixed feature linear evaluation on the 12 downstream datasets for models from different pretraining epochs. The transfer performance of the \CE pretrained model improves very little after 80 pretraining epochs, suggesting that \CE learns transferable representation mostly during the initial phase of the pretraining, and it learns more source-domain specific representation at later pretraining epochs. On the other hand, transferability of contrastive methods improve gradually with longer source dataset pretraining. Moreover, during the initial pretraining stages, we find that both \CEplusMOCO and \MITplusMOCO still perform better in transfer learning than other models, suggesting that the joint objectives could be beneficial even in resource constrained environments.  

We refer the reader to the Appendix (included in the supplementary material) for more results and discussions.

\section{Conclusion}
\label{sec:conclusion}

In this paper, we conduct extensive analysis on the transferability of contrastive learning on the downstream image classification, few-shot recognition, and object detection tasks. Our study suggests that contrastive models consistently perform better in transfer learning than standard cross-entropy models, and combining self-supervised contrastive loss with cross-entropy or supervised contrastive loss improves transfer learning performance. We find several factors that make representations from contrastive learning more transferable than supervised cross-entropy model.  The penultimate layer representations of contrastive learning are much different than the cross-entropy model;  in particular, contrastive models contain more low-level and mid-level information in final layers, and the contrastively trained model shows larger intra-class separation, and contrastive models are more robust to image corruptions. 


\section{Acknowledgement}
This material is based upon work supported by the Defense Advanced Research Projects Agency (DARPA) under Contract No. FA8750-19-C-1001. Any opinions, findings and conclusions or recommendations expressed in this material are those of the author(s) and do not necessarily reflect the views of the Defense Advanced Research Projects Agency (DARPA).

{\small
\bibliographystyle{ieee_fullname}
\bibliography{main}
}

\newpage
\clearpage
\appendix
The supplemental material contains additional analysis, visualization and ablation studies.  All these are not included in the main paper due to the space limit.

\section{Experimental Details} \label{ap:exp_details}
\subsection{Datasets} \label{ap:dataset}
For the source dataset training, we use ImageNet1K training set~\cite{deng2009imagenet}. For downstream linear evaluation, we use 12 datasets from different domains to evaluate the transferability of different models. We divide the datasets into six groups - natural, satellite, symbolic, illustrative, medical, and texture. Unless otherwise mentioned, we use top-1 accuracy as evaluation metrics. 

The most similar to ImageNet categories (i.e., natural images) are CropDisease, DeepWeeds and Flowers102.  \textbf{CropDisease}~\cite{mohanty2016cropdisease} contains natural images of diseased crop leaves categorized into 38 different classes. ~\textbf{DeepWeeds}~\cite{DeepWeeds2019} contains 17,509 images of 8 different weed species native to Australia. \textbf{Flowers}~\cite{nilsback2008automatedflowers102} is a fine-grained dataset of 102 different flower categories each of which consists of 40 to 258 images.      

In the satellite image category, we use EuroSAT adn Resisc45. \textbf{EuroSAT}~\cite{helber2019eurosat} is a satellite imagery dataset consisting of 27,000 labeled images with 10 different land use and land cover classes.  \textbf{Resisc}~\cite{cheng2017remoteresisc45} is a remote sensing image classification dataset containing 31,500 images of 45 scene classes. 

\textbf{SVHN}~\cite{netzer2011readingsvhn} is obtained from house numbers of google street view images. There are 10 classes, 1 for each digit from 0 to 9, consisting of around 73k training images and 26k testing images. \textbf{Omniglot} \cite{lake2015humanomniglot} contains 1623 different hand-writted characters from 50 different alphabets. 

Both Kaokore and Sketch contain illustrative or hand-drawn images. \textbf{Kaokore}~\cite{tian2020kaokore} dataset contains 8848 face images from japanese illustration. ImageNet ~\textbf{Sketch}~\cite{wang2019learningsketch} consists of `black-and-white' sketches from each of ImageNet 1000 classes. Kaokore dataset contains two super classes based on gender and status. The `gender' class contains male and female, and the `status' class contains - noble, warrior, incarnation, and commoner. Combining both of them we get total 8 classes.

In the medical imagery domain, we have ChestX and ISIC dataset. \textbf{ChestX} \cite{wang2017chestx} is comprised of X-Ray images, and \textbf{ISIC}~\cite{codella2019skinisic} dataset contains dermoscopic images of skin lesions. 

In the category of texture dataset, we use \textbf{DTD}~\cite{cimpoi2014DTD}, which consists of 5640 texture images from 47 categories.  
We use the official training and test split for most datasets if available. If the official split contains both training and validation split, we combine them for training. When there are multiple official splits available (e.g. DTD \cite{cimpoi2014DTD}), we use the first split for evaluation. If there is no official split available, we randomly select 30\% of the total images from each category as test set and the remaining images for training. For few-shot learning, we use all the available images from both training and test splits, and randomly select images for support set and query set from 5 random classes (5-way few-shot learning) at each few-shot episode.

\subsection{ImageNet Pretraining} \label{ap:in_pretrain}
We trained all models on the source dataset for 400 epochs with learning rate $0.01$ and cosine scheduling with warm-up for 5 epochs. 
We set the temperature parameter of MoCo to $\tau=0.07$, and queue size to $65,596$ \cite{he2020momentummoco}. The effect of queue size to the transfer learning performance is discussed in the Appendix \ref{ap:queue_size}. For \CE model, we use random crop(224x224), horizontal flip, and normalization data-augmentations during training. For \MOCO, \MITplusMOCO, and \CEplusMOCO, we use random-crop (224x224), color-jitter, random gray-scale, Gaussian blur, random horizontal flip, and normalization for training data augmentations. CE models with stronger augmentation is also discussed in Appendix.~\ref{ap:ce_strong_abl}. Unless otherwise mentioned, we use top-1 accuracy as the evaluation metric. 

\subsection{Linear Evaluation} \label{ap:linear_eval}
For fixed-feature linear evaluation, we freeze the pretrained backbone, add a linear layer to train it on the downstream dataset. We add a BatchNorm layer without any affine parameter between the backbone and linear layer to make the extracted features comparable among different models. Note that the BatchNorm layer makes the models to have similar optimal hyperparameters during linear evaluation. We train all models for 50 epochs with step learning rate scheduler which decreases the learning rate by 0.1 at epoch 25 and 37. We also experimented with 100 epochs for all models, but did not notice any noticeable improvement over 50 epochs. As different datasets might require different hyperparameters, we perform extensive hyperparameter tuning. We split the training set on 70\% training and 30\% validation, and then train the models for
\begin{itemize}
    \item learning rate: 0.001, 0.01, 0.1
    \item batch-size: 32, 128
    \item weight decay: 0, 1e-4, 1e-5
\end{itemize}
and chose the optimal hyperparamters among different runs  based on the performance on the validation set. Nevertheless, we found that batch-size 128, learning rate 0.01, and weight-decay 0 can be chosen as a safe hyperparameter choice for most cases.

\subsection{Object Detection}
For object detection, we follow the setting in \cite{he2020momentummoco} to finetune the full network, and add batch normalization layer into the meta architecture of the detector, e.g., region-of-interested (ROI) header, feature pyramid network (FPN), etc., to minimize the effort on hyperparameter tuning.

\subsection{Full-network Finetuning} \label{ap:full_tune}
We train the whole network, i.e., the pretrained backbone and linear layer on the downstream dataset. We train all models for 50 epochs with step learning rate scheduler which decreases the learning rate by 0.1 at epoch 25 and 37. Here, we also experimented with 100 epochs, but did not notice any noticeable improvement over 50 epochs. We add a BatchNorm layer between the backbone and linear layer to make the extracted features comparable among different models. We train the models for various learning rate - 0.01, 0.001, and 0.1, batch-sizes - 32 and 128, and weight decay 0, 1e-4, 1e-5 and select the optimal hyperparameter based on the performance on the validation set. We found that for most datasets learning rate 0.001 with batch-size 32 performs the best, hence this setting can be used in a scenario where hyperperameter tuning is not possible or expensive.  

\subsection{Few-shot Recogniton} \label{ap:fewshot_setup}

For few-shot pretraining, we closely followed CDFSL benchmark paper \cite{guo2020broaderstudy}. For all models, we train for 300 epochs with SGD optimizer with learning rate 0.01 and batch-size 32, and cosine scheduling with warm-up for 5 epochs. For the contrastive models, we use number of negative samples to be 16384 and temperature parameter to be 0.07.  Following \cite{tian2020rethinkingfewshotgoodembedding}, we train a logistic regression layer on top of the extracted features during meta-testing phase. We use the implementation from scikit-learn for logistic regression \cite{scikitlearn}. The accuracy is the mean of 600 randomly sampled tasks, and 95\% confidence interval is also reported.

\section{More Results and Analysis} \label{ap:more_results_ab}

\subsection{Linear Evaluation} \label{ap:linear_eval_res}
Detailed numbers for linear evaluation are provided in \ref{tab:linbn_score_app}. 

\begin{table*}[!htbp]
    \centering
    \begin{adjustbox}{max width=\linewidth}
    \begin{tabular}{l|llllllllllll|l}
\toprule
{} &             \rot{CropDisease} &               \rot{DeepWeeds} &              \rot{Flowers102} &                 \rot{EuroSAT} &                \rot{Resisc45} &                    \rot{ISIC} &                  \rot{ChestX} &                \rot{Omniglot} &                    \rot{SVHN} &                  \rot{Kaokore} &                  \rot{Sketch} &                     \rot{DTD} &                    \rot{Mean} \\
\midrule
\CE          &  97.18{\small $\pm$.04} &  85.17{\small $\pm$.04} &  84.73{\small $\pm$.52} &  94.62{\small $\pm$.12} &  84.96{\small $\pm$.25} &  79.58{\small $\pm$.21} &  44.70{\small $\pm$.20} &  63.88{\small $\pm$.61} &  66.31{\small $\pm$.16} &   73.96{\small $\pm$.16} &  67.41{\small $\pm$.13} &  65.48{\small $\pm$.40} &  75.67{\small $\pm$.24} \\
\MOCO        &  99.06{\small $\pm$.04} &  87.88{\small $\pm$.16} &  89.62{\small $\pm$.14} &  96.76{\small $\pm$.05} &  90.88{\small $\pm$.10} &  81.51{\small $\pm$.35} &  48.08{\small $\pm$.13} &  69.66{\small $\pm$.23} &  69.95{\small $\pm$.06} &   81.67{\small $\pm$.80} &  69.12{\small $\pm$.13} &  72.21{\small $\pm$.57} &  79.70{\small $\pm$.23} \\
\MIT         &  98.85{\small $\pm$.03} &  87.37{\small $\pm$.13} &  92.79{\small $\pm$.15} &  96.05{\small $\pm$.12} &  90.26{\small $\pm$.24} &  79.78{\small $\pm$.22} &  46.76{\small $\pm$.32} &  72.99{\small $\pm$.31} &  74.09{\small $\pm$.12} &   80.94{\small $\pm$.23} &  77.19{\small $\pm$.10} &  74.00{\small $\pm$.37} &  80.92{\small $\pm$.20} \\
\bottomrule
\end{tabular}
    \end{adjustbox}
    \caption{\small Top-1 accuracy of different models on the downstream datasets for fixed-feature extractor transfer learning. The models are pretrained on ImageNet1K dataset and we only train the final linear layer on top of the pretrained backbones. Mean and standard deviation over 5-runs are provided.}
    \label{tab:linbn_score_app}
\end{table*}

\subsection{Object Detection}

Table \ref{tab:obj_detection_details_app} shows object detection results with standard deviation over 5 runs.
\begin{table*}[!thbp]
    \centering
    \begin{adjustbox}{max width=\linewidth}
        \begin{tabular}{l| ccc || c c c | c c c || c c c | c c c}
    \toprule
        Datasets & \multicolumn{3}{c||}{VOC0712} & \multicolumn{12}{c}{MS COCO (Trained with 1$\times$ schedule)} \\ 
        Detectors & \multicolumn{3}{c||}{FasterRCNN-R50-C4}  & \multicolumn{6}{c||}{MaskRCNN-R50-C4}  & \multicolumn{6}{c}{MaskRCNN-R50-FPN} \\ 
        Methods &  AP$^{\text{bb}}_\text{50}$ &  AP$^{\text{bb}}$ & AP$^{\text{bb}}_{\text{75}}$ &  AP$^{\text{bb}}$ &  AP$^{\text{bb}}_{\text{50}}$ & AP$^{\text{bb}}_{\text{75}}$ &  AP$^{\text{mk}}$ &  AP$^{\text{mk}}_{\text{50}}$ & AP$^{\text{mk}}_{\text{75}}$ &  AP$^{\text{bb}}$ &  AP$^{\text{bb}}_{\text{50}}$ & AP$^{\text{bb}}_{\text{75}}$ &  AP$^{\text{mk}}$ &  AP$^{\text{mk}}_{\text{50}}$ & AP$^{\text{mk}}_{\text{75}}$\\
    \midrule
    

\CE & 81.58{\small $\pm$.17}	&	54.63{\small $\pm$.28}	&	60.17{\small $\pm$.40}	&	38.91{\small $\pm$.02}	&	59.05{\small $\pm$.20}	&	42.14{\small $\pm$.16}	&	33.98{\small $\pm$.04}	&	55.39{\small $\pm$.14}	&	36.10{\small $\pm$.12}	&	39.84{\small $\pm$.14}	&	60.54{\small $\pm$.13}	&	43.60{\small $\pm$.22}	&	36.54{\small $\pm$.08}	&	57.65{\small $\pm$.10}	&	39.25{\small $\pm$.18} \\
\end{tabular}
    
    \end{adjustbox}
    \caption{Object detection results on MS COCO.}
    \label{tab:obj_detection_details_app}
\end{table*}

\begin{table*}[!thbp]
    \centering
    \begin{adjustbox}{max width=\linewidth}
    \begin{tabular}{l| c c c | c c c || c c c | c c c}
    \toprule
        Detectors &  \multicolumn{6}{c||}{MaskRCNN-R50-C4}  & \multicolumn{6}{c}{MaskRCNN-R50-FPN} \\ 
        Methods &   AP$^{\text{bb}}$ &  AP$^{\text{bb}}_{\text{50}}$ & AP$^{\text{bb}}_{\text{75}}$ &  AP$^{\text{mk}}$ &  AP$^{\text{mk}}_{\text{50}}$ & AP$^{\text{mk}}_{\text{75}}$ &  AP$^{\text{bb}}$ &  AP$^{\text{bb}}_{\text{50}}$ & AP$^{\text{bb}}_{\text{75}}$ &  AP$^{\text{mk}}$ &  AP$^{\text{mk}}_{\text{50}}$ & AP$^{\text{mk}}_{\text{75}}$\\
    \midrule

        \CE & 41.04	&	60.86	&	44.63	&	35.48	&	57.29	&	37.82	&	41.43	&	62.12	&	45.38	&	37.89	&	59.29	&	40.53 \\ 
\MOCO & 41.00	&	60.77	&	44.43	&	35.68	&	57.48	&	38.21	&	41.15	&	61.53	&	45.05	&	37.63	&	58.84	&	40.59 \\ 
\MIT & 41.58	&	61.72	&	44.99	&	36.04	&	58.16	&	38.42	&	41.70	&	62.39	&	45.47	&	38.03	&	59.72	&	40.66 \\ 
\CEplusMOCO & 41.28	&	61.26	&	45.06	&	35.75	&	57.60	&	38.34	&	41.77	&	62.33	&	45.78	&	38.15	&	59.63	&	41.10 \\ 
\MITplusMOCO & 41.28	&	61.18	&	44.97	&	35.91	&	57.72	&	38.38	&	41.96	&	62.58	&	45.92	&	38.37	&	59.81	&	41.36 \\
\end{tabular}
    \end{adjustbox}
    \caption{Object detection and instance segmentation results on MS COCO (2$\times$ schedule).}
    \label{tab:od_2x}
\end{table*}

\begin{table}[!thbp]
    \centering
    \begin{adjustbox}{max width=\linewidth}
    \begin{tabular}{l| c c c | c c c }
    \toprule
        Detector &  \multicolumn{6}{c}{RetinaNet-R50}  \\ 
            &   \multicolumn{3}{c|}{1$\times$ schedule} & \multicolumn{3}{c}{2$\times$ schedule}\\
        Methods &   AP$^{\text{bb}}$ &  AP$^{\text{bb}}_{\text{50}}$ & AP$^{\text{bb}}_{\text{75}}$  &   AP$^{\text{bb}}$ &  AP$^{\text{bb}}_{\text{50}}$ & AP$^{\text{bb}}_{\text{75}}$ \\
    \midrule
        \CE             & 38.46	&	58.32	&	41.19	&	40.13	&	60.02	&	43.10 \\ 
        \MOCO           & 37.65	&	56.62	&	40.29	&	39.61	&	59.14	&	42.28 \\ 
        \MIT            & 38.77	&	58.48	&	41.52	&	40.35	&	60.23	&	43.26 \\ 
        \CEplusMOCO     & \textbf{39.33}	&	58.94	&	42.24	&	40.52	&	60.11	&	43.32 \\ 
        \MITplusMOCO    & 39.24	&	58.67	&	42.23	&	40.50	&	60.34	&	43.50 \\

    \bottomrule
    \multicolumn{7}{l}{\footnotesize{AP$^{\text{bb}}$: AP of objection detection.}} \\
    \end{tabular}
    \end{adjustbox}
    \caption{Object detection results on MS COCO.}
    \label{tab:od_retina}
\end{table}

We also conducted the experiments of object detection with longer training setting (2$\times$ schedule in Detectron2.) and tested another detector, RetinaNet-R50. The results are shown in Table \ref{tab:od_2x} and Table \ref{tab:od_retina}.
When training with more iterations (2$\times$ schedule), the results of \CE, \MOCO and \MIT are more closer to \CEplusMOCO and \MITplusMOCO as pretraining between less important if the downstream has been trained for a long time. The results of the RetinaNet have the similar trend, which shows that the visual representations trained by \CEplusMOCO and \MITplusMOCO are transferable on different types of detectors.

\subsection{Few-shot Recogniton} \label{ap:fewshot_results}
5-way 1-shot, 5-shot and 20-shot results are provided in Table \ref{tab:fewshot_app}. 

\begin{table*}[!thbp]
    \centering
    \begin{adjustbox}{max width=\linewidth}
    \begin{tabular}{ll||llllllllllll|l}
\toprule
{} &            \rot{miniImagenet} &             \rot{CropDisease} &               \rot{DeepWeeds} &              \rot{Flowers102} &                 \rot{EuroSAT} &                \rot{Resisc45} &                    \rot{ISIC} &                  \rot{ChestX} &                \rot{Omniglot} &                    \rot{SVHN} &                 \rot{Kaokore} &                  \rot{Sketch} &                     \rot{DTD} &                    \rot{Mean} \\
\midrule
\multicolumn{14}{l}{1-shot} \\
\midrule
\CE          &  53.96{\small $\pm$.82} &  68.72{\small $\pm$.85} &  35.73{\small $\pm$.62} &  62.03{\small $\pm$.92} &  63.47{\small $\pm$.91} &  53.38{\small $\pm$.75} &  33.33{\small $\pm$.60} &  22.78{\small $\pm$.41} &  83.73{\small $\pm$.71} &  23.37{\small $\pm$.41} &  28.92{\small $\pm$.54} &  45.25{\small $\pm$.79} &  41.43{\small $\pm$.75} &  46.84{\small $\pm$.69} \\
\MOCO        &  48.88{\small $\pm$.77} &  65.35{\small $\pm$.85} &  38.68{\small $\pm$.61} &  66.04{\small $\pm$.85} &  66.07{\small $\pm$.83} &  53.09{\small $\pm$.86} &  33.73{\small $\pm$.60} &  22.93{\small $\pm$.43} &  81.88{\small $\pm$.74} &  23.28{\small $\pm$.40} &  31.70{\small $\pm$.61} &  47.83{\small $\pm$.79} &  44.19{\small $\pm$.79} &  47.90{\small $\pm$.70} \\
\MIT         &  55.50{\small $\pm$.78} &  63.59{\small $\pm$.83} &  35.71{\small $\pm$.59} &  63.17{\small $\pm$.86} &  63.41{\small $\pm$.83} &  55.14{\small $\pm$.81} &  32.15{\small $\pm$.60} &  22.73{\small $\pm$.41} &  87.52{\small $\pm$.66} &  25.82{\small $\pm$.46} &  32.87{\small $\pm$.59} &  53.55{\small $\pm$.85} &  48.22{\small $\pm$.82} &  48.66{\small $\pm$.69} \\
\bottomrule
\end{tabular}
    \end{adjustbox}
    \caption{\small Few-shot classification accuracies (\%) on the Mini-ImageNet and 12 downstream datasets. Mean and 95\% confidence interval over 600 tasks.}
    \label{tab:fewshot_app}
\end{table*}

\subsection{Full-network Finetune} \label{ap:ftune_results}
Detailed results for image classification with full-network finetuning are provided in Table \ref{tab:tfmbn_app}.

\begin{table*}[!thbp]
    \centering
    \begin{adjustbox}{max width=\linewidth}
    \begin{tabular}{l|llllllllllll|l}
\toprule
{} &             \rot{CropDisease} &               \rot{DeepWeeds} &               \rot{Flowers102} &                 \rot{EuroSAT} &                \rot{Resisc45} &                    \rot{ISIC} &                  \rot{ChestX} &                \rot{Omniglot} &                    \rot{SVHN} &                 \rot{Kaokore} &                  \rot{Sketch} &                     \rot{DTD} &                    \rot{Mean} \\
\midrule
\multicolumn{14}{l}{Full Dataset} \\
\midrule
\CE          &  99.92{\small $\pm$.02} &  97.37{\small $\pm$.13} &   93.47{\small $\pm$.30} &  98.78{\small $\pm$.09} &  96.03{\small $\pm$.09} &  88.06{\small $\pm$.36} &  55.67{\small $\pm$.56} &  90.34{\small $\pm$.29} &  97.03{\small $\pm$.12} &  88.13{\small $\pm$.40} &  79.49{\small $\pm$.27} &  73.26{\small $\pm$.56} &  88.13{\small $\pm$.27} \\
\MOCO        &  99.91{\small $\pm$.02} &  97.39{\small $\pm$.13} &  93.36{\small $\pm$2.07} &  98.85{\small $\pm$.04} &  96.28{\small $\pm$.12} &  88.13{\small $\pm$.13} &  56.41{\small $\pm$.26} &  91.10{\small $\pm$.20} &  97.25{\small $\pm$.10} &  88.92{\small $\pm$.36} &  77.16{\small $\pm$.13} &  75.49{\small $\pm$.17} &  88.35{\small $\pm$.31} \\
\MIT         &  99.91{\small $\pm$.01} &  96.89{\small $\pm$.11} &   95.37{\small $\pm$.16} &  98.67{\small $\pm$.12} &  96.01{\small $\pm$.06} &  87.92{\small $\pm$.33} &  55.08{\small $\pm$.57} &  90.56{\small $\pm$.19} &  97.11{\small $\pm$.11} &  87.88{\small $\pm$.66} &  80.14{\small $\pm$.19} &  74.19{\small $\pm$.20} &  88.31{\small $\pm$.23} \\
\bottomrule
\end{tabular}
    \end{adjustbox}
    \caption{\small  Performance of different models on the downstream datasets in terms of top-1 accuracy (\%) (averaged over 5 runs) for full-network fine-tuning. Contrastive pretrained methods are slightly more effective in a limited data regime than cross-entropy based models.}
    \label{tab:tfmbn_app}
\end{table*}

\section{More Analysis} \label{ap:more_analysis}

\subsection{Experimental Setup for Analysis on ImageNet Derivatives} \label{ap:in_derivative}

For the evaluation on ImageNet1K, ImageNet-A, ImageNet-R, ImageNet-C, and Stylized ImageNet on Table \ref{tab:res_on_IN_val} and Table \ref{tab:robust_main} in the main paper, we train a classifier header on top of the ResNet-50 backbone for the contrastive models. The classifier header is trained on the ImageNet1K training set. We SGD optimizer, and tune the learning rate for 0.01, 0.1, 1, 2, 10, and chose the best performing model based on the validation set accuracy. We do not use any weight decay \cite{he2020momentummoco}. Note that this is performed only for \MOCO, \MIT, \MITplusMOCO. For \CE and \CEplusMOCO, we use the header associated with the supervised branch. 


\subsection{Model Calibration on the Downstream Datasets} \label{ap:calib_downstream}
Table \ref{tab:calibration_transfer_app} reports calibration performance for the ImageNet pretrained models on the 12 downstream datasets in terms of Expected Calibration Error (ECE). We use fixed-feature linear evaluation to train the linear layer on the downstream tasks. We also perform hyperparameter sweaping on the validation set and report the best ECE score (lower is better). We observe that contrastive models have lower calibration error than cross-entropy model on average.

\begin{table*}[!htbp]
    \centering
    \begin{adjustbox}{max width=\linewidth}
    \begin{tabular}{l|llllllllllll|l}
\toprule
{} &            \rot{CropDisease} &              \rot{DeepWeeds} &             \rot{Flowers102} &                \rot{EuroSAT} &               \rot{Resisc45} &                   \rot{ISIC} &                 \rot{ChestX} &                \rot{Omniglot} &                   \rot{SVHN} &                  \rot{Kaokore} &                  \rot{Sketch} &                     \rot{DTD} &                   \rot{Mean} \\
\midrule
\CE          &  0.83{\small $\pm$.06} &  1.41{\small $\pm$.37} &  2.98{\small $\pm$.33} &  0.64{\small $\pm$.11} &  6.89{\small $\pm$.32} &  1.74{\small $\pm$.26} &  2.79{\small $\pm$.26} &   9.30{\small $\pm$.43} &  2.10{\small $\pm$.28} &    4.84{\small $\pm$.20} &  9.60{\small $\pm$1.45} &   6.64{\small $\pm$.55} &  4.15{\small $\pm$.38} \\
\MOCO        &  0.24{\small $\pm$.03} &  3.58{\small $\pm$.15} &  2.52{\small $\pm$.12} &  0.78{\small $\pm$.08} &  1.02{\small $\pm$.21} &  2.92{\small $\pm$.45} &  2.01{\small $\pm$.47} &  19.02{\small $\pm$.66} &  3.99{\small $\pm$.23} &    3.93{\small $\pm$.64} &   6.46{\small $\pm$.18} &   7.08{\small $\pm$.56} &  4.46{\small $\pm$.31} \\
\MIT         &  0.55{\small $\pm$.03} &  1.80{\small $\pm$.15} &  4.21{\small $\pm$.12} &  0.99{\small $\pm$.15} &  4.83{\small $\pm$.21} &  2.05{\small $\pm$.20} &  1.80{\small $\pm$.25} &   3.82{\small $\pm$.33} &  2.52{\small $\pm$.25} &    3.37{\small $\pm$.54} &   7.27{\small $\pm$.23} &   3.53{\small $\pm$.24} &  3.06{\small $\pm$.23} \\
\bottomrule
\end{tabular}
    \end{adjustbox}
    \caption{\small Expected calibration Error (\%) of the models on the downstream datasets. The models are pre-trained on ImageNet1K dataset and we only train the final linear layer on top of the pretrained backbones.}
    \label{tab:calibration_transfer_app}
\end{table*}

\subsection{Robustness of Adversarial Attack}
\begin{figure}[!thbp]
    \centering
    \includegraphics[width=\linewidth]{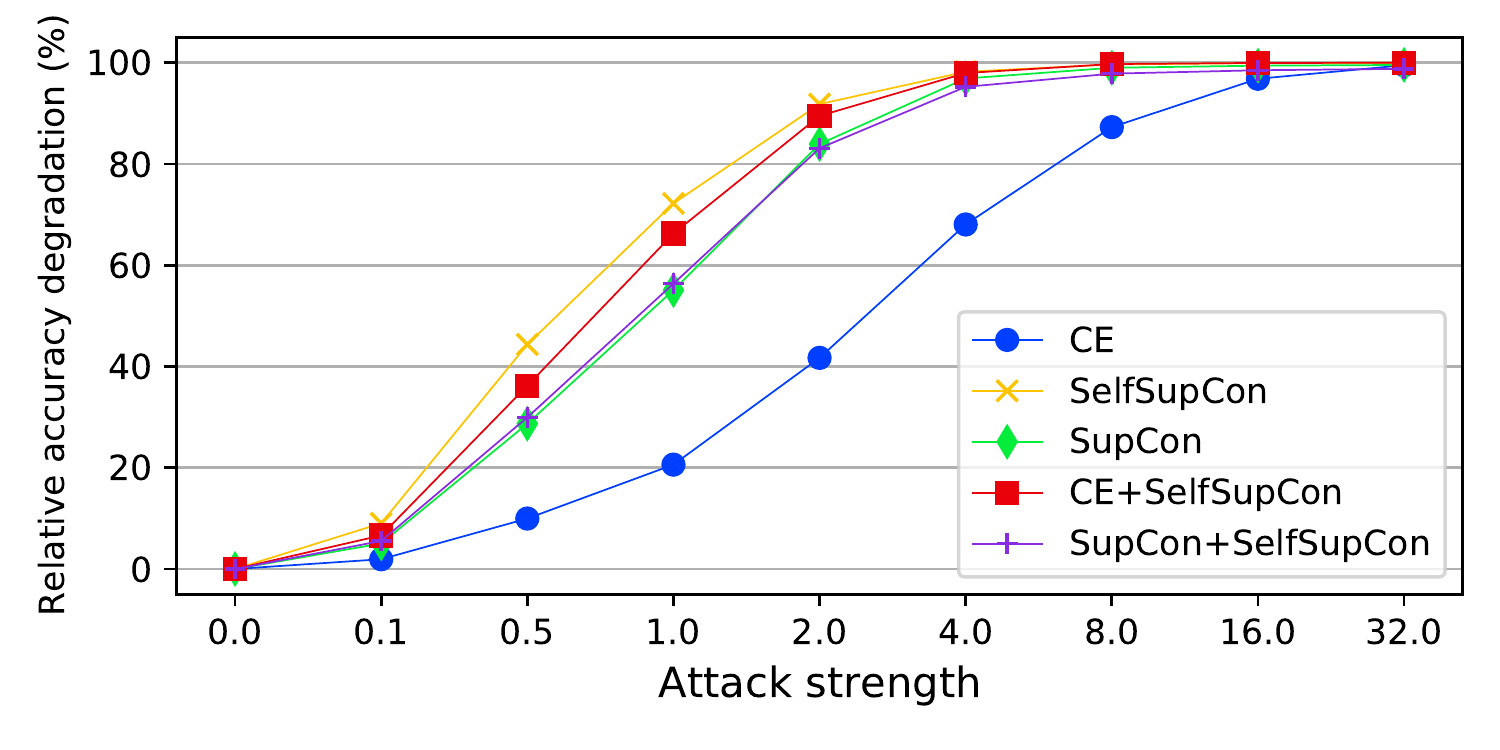} 
    \caption{\small{Performance degradation under different attack strengths ($\varepsilon$). $L_{\infty}$PGD attack is used~\cite{madry2018LinfPGD}.}}
    \label{fig:adv_attack}
\end{figure}

We used projected gradient descent (PGD) attack to test the robustness of each model. It is clear that without any adversarial training, all models are vulnerable. Nonetheless, we would like to analyze the region with smaller perturbation (i.e., small $\varepsilon$). Figure~\ref{fig:adv_attack} show the relative accuracy degradation among those methods. Interestingly, \CE is the most robust one among while all other methods are relatively vulnerable. We think it is because contrastive learning methods are more sensitive to the local changes as they learn local features while \CE does not; hence, \CE can tolerance more perturbations.

\section{More Ablation Studies} \label{ap:more_ablation}

\subsection{Effect of Queue-size to Transferability} \label{ap:queue_size}
Table \ref{tab:queue_effect_app} shows the effect of queue size to transferability for the contrastive models. All models are trained on ImageNet1K training set with queue sizes 1024, 8192, and 65596. We perform fixed-feature linear evaluation on the downstream datasets and report the average accuracy. The table shows that higher queue size is better for transferability. 
 
\begin{table}[!htbp]
    \centering
    \begin{adjustbox}{max width=\linewidth}
    \begin{tabular}{l|c c c}
\toprule
\multirow{2}{*}{Method} & \multicolumn{3}{c}{Queue size} \\
& 1024 & 8192 & 65596 \\
\midrule
    \MOCO & 79.23 & 79.68 & 79.70 \\
    \MIT & 78.11 & 79.88 & 80.92  \\
    \CEplusMOCO &  77.24 & 78.66 & 80.19 \\
    \MITplusMOCO &  79.42 & 80.50 & 81.30 \\
\bottomrule
\end{tabular}

    \end{adjustbox}
    \caption{Ablation studies on the effect of \textbf{queue size} to the linear evaluation performance on the downstream datasets. Models trained with higher queue size generally performs better in transfer. }
    \label{tab:queue_effect_app}
\end{table}

\subsection{Results with torchvision-pretrained Model} \label{ap:torchvision}

Table \ref{tab:linbn_score_app_torch} shows linear evaluation and full-network finetuning scores from torchvision pretrained model and our reconstructed ResNet-50 model. We note that the torchvision pretrained one achieves better accuracy in linear evaluation, but the accuracy is similar in full-network finetuning. Our \CE model was trained for 400 epochs with cosine learning rate scheduler, whereas the torchvision one has been trained for 100 epochs with step learning scheduler. We hypothesize that the training schedule might have made difference in the transfer performance. However, although the torchvision pretrained model has better linear evaluation accuracy, it is not better in full-network finetuning, robustness or even ImageNet performance. The \CE (torchvision) achieves 75.7\% ImageNet accuracy, where our \CE achieves 76.60\% ImageNet accuracy. From Table \ref{tab:robust_app}, our \CE achieves 60.80\% mCE in ImageNet-C, whereas \CE (torchvision) achieves much worse score 78.50\%. We infer that our training strategy might be better for cross-entropy model in terms of calibration, robustness, or even in-domain accuracy. Overall, even with the torchvision pretrained ResNet-50 model, our main thesis does not change, as the performance is still significantly lower than the contrastive models.

\begin{table*}[!htbp]
\centering
\begin{adjustbox}{max width=\linewidth}
\begin{tabular}{l|llllllllllll|l}
\toprule
{} &            \rot{CropDisease} &               \rot{DeepWeeds} &              \rot{Flowers102} &                 \rot{EuroSAT} &                \rot{Resisc45} &                    \rot{ISIC} &                  \rot{ChestX} &                \rot{Omniglot} &                    \rot{SVHN} &                 \rot{Kaokore} &                  \rot{Sketch} &                     \rot{DTD} &                    \rot{Mean} \\
\midrule
Linear Evaluation &  98.19{\small $\pm$.05} &  87.58{\small $\pm$.18} &  88.40{\small $\pm$.18} &  95.21{\small $\pm$.03} &  85.62{\small $\pm$.06} &  76.87{\small $\pm$.13} &  45.05{\small $\pm$.23} &  64.39{\small $\pm$.34} &  70.41{\small $\pm$.11} &  75.79{\small $\pm$.11} &  66.18{\small $\pm$.43} &  71.72{\small $\pm$.24} &  77.12{\small $\pm$.17} \\
Finetune &  99.87{\small $\pm$.01} &  97.53{\small $\pm$.09} &  95.47{\small $\pm$.20} &  98.91{\small $\pm$.05} &  95.92{\small $\pm$.19} &  88.74{\small $\pm$.27} &  54.47{\small $\pm$.35} &  89.99{\small $\pm$.14} &  96.96{\small $\pm$.04} &  87.73{\small $\pm$.30} &  78.36{\small $\pm$.14} &  73.84{\small $\pm$.27} &  88.15{\small $\pm$.17} \\
\bottomrule
\end{tabular}
    \end{adjustbox}
    \caption{\small Top-1 accuracy of different models on the downstream datasets for ImageNet-pretrained ResNet-50 model from \textbf{torchvision} \cite{NEURIPS2019_9015pytorch}. Mean and std over 5 runs.}
    \label{tab:linbn_score_app_torch}
\end{table*}

\begin{table}[htb!]
    \centering
    \begin{adjustbox}{max width=\linewidth}
    \begin{tabular}{l|c c | c c| c}
\toprule
    \multirow{2}{*}{Method} & \multicolumn{2}{c|}{ImageNet-R} & \multicolumn{2}{c|}{ImageNet-A} & ImageNet-C \\ 
    & Top-1(\%) & ECE(\%) & Top-1(\%) & ECE(\%) & mCE(\%) \\
\midrule
    \CE &  35.83 & 19.45 & 3.35 & 55.03 & 60.80 \\
    \CE (torchvision) & 36.11 & 19.75 &  3.98 & 62.44  & 78.50 \\
    \CEstrong &  41.24 & 17.90 & 8.01 & 50.81 & 68.90 \\
\bottomrule
\end{tabular}
    \end{adjustbox}
    \caption{\small Robustness tests on ImageNet-R, ImageNet-A, and ImageNet-C datasets for \CEstrong and \CE (torchvision). ECE is the expected calibration error (lower is better) and mCE (lower is better) is the mean of the (unnormalized) corruption errors of the Noise, Blur, Weather, and Digital corruptions. Models are trained only on clean ImageNet images.}
    \label{tab:robust_app}
\end{table}

\subsection{ResNet50x2 as Backbone} \label{ap:wide_backbone}
We also use WideResNet-50-x2 as the backbone of the networks, and report fixed feature linear evaluation transfer and  full-network transfer in Table \ref{tab:linear_wres_app} and. We found similar pattern with larger backbone that contrastive approaches provide more transferable representations. 

\begin{table*}[!htbp]
    \centering
    \begin{adjustbox}{max width=\linewidth}
    \begin{tabular}{l|rrrrrrrrrrrr|r}
\toprule
{} &  \rot{CropDisease} &  \rot{DeepWeeds} &  \rot{Flowers102} &  \rot{EuroSAT} &  \rot{Resisc45} &  \rot{ISIC} &  \rot{ChestX} &  \rot{Omniglot} &  \rot{SVHN} &  \rot{Kaokore} &  \rot{Sketch} &   \rot{DTD} &  \rot{Mean} \\
\midrule
\multicolumn{14}{l}{Linear Evaluation} \\
\midrule
\CE          &        97.27 &      87.33 &       91.08 &    95.04 &     85.41 & 80.39 &   45.50 &     64.90 & 67.85 &    77.59 &   69.14 & 68.40 & 77.49 \\
\MOCO        &        99.07 &      88.23 &       92.45 &    97.10 &     90.63 & 81.62 &   49.96 &     71.72 & 71.57 &    82.10 &   69.28 & 71.33 & 80.42 \\
\MIT         &        98.52 &      86.70 &       94.90 &    95.99 &     90.55 & 79.32 &   47.62 &     70.26 & 74.76 &    81.24 &   76.62 & 74.36 & 80.90 \\
\CEplusMOCO  &        98.82 &      89.48 &       94.51 &    96.51 &     89.94 & 80.22 &   49.56 &     69.40 & 73.34 &    79.42 &   75.08 & 74.41 & 80.89 \\
\MITplusMOCO &        98.95 &      89.21 &       96.27 &    96.95 &     91.50 & 80.35 &   49.46 &     73.52 & 73.11 &    81.61 &   76.24 & 74.89 & 81.84 \\

\midrule
\multicolumn{13}{l}{Full-network finetuning} \\
\midrule

\CE          &        99.92 &      97.66 &       97.75 &    98.75 &     96.05 & 89.30 &   55.54 &     91.02 & 96.90 &    90.01 &   80.42 & 73.35 & 88.89 \\
\MOCO        &        99.88 &      96.94 &       97.84 &    98.81 &     96.17 & 89.56 &   56.70 &     91.15 & 97.13 &    89.52 &   77.08 & 74.52 & 88.78 \\
\MIT         &        99.86 &      96.99 &       98.63 &    98.62 &     95.89 & 88.76 &   55.68 &     90.77 & 97.00 &    89.52 &   80.98 & 75.32 & 89.00 \\
\CEplusMOCO  &        99.86 &      97.24 &       98.24 &    98.78 &     96.12 & 90.09 &   55.12 &     90.97 & 97.07 &    90.01 &   81.94 & 77.23 & 89.39 \\
\MITplusMOCO &        99.89 &      97.60 &       97.94 &    98.74 &     96.29 & 89.96 &   55.83 &     91.05 & 97.05 &    90.74 &   81.69 & 76.60 & 89.45 \\

\bottomrule
\end{tabular}
    \end{adjustbox}
    \caption{Performance of different models with \textbf{WideResNet-50-x2} backbone on the downstream datasets in terms of top-1 accuracy with fixed-feature \textbf{linear evaluation}. We present the best performing scores for different hyperparameters for each model.}
    \label{tab:linear_wres_app}
\end{table*}


\subsection{Pretraining on Stylized ImageNet} \label{ap:pretrain_SIN}

We also train all models on the Stylized ImageNet training set so that the models learn more shape-based representations \cite{geirhos2018SIN}. We then perform a fixed feature linear evaluation on the 12 downstream datasets. Table \ref{tab:linear_SIN_app} shows the top-1 accuracy on the downstream datasets for the Stylized ImageNet trained model. The results reveal that contrastive approaches also provide better transferable representation than the cross-entropy model when trained on Stylized ImageNet. 

\begin{table*}[!htbp]
    \centering
    \begin{adjustbox}{max width=\linewidth}
    \begin{tabular}{l|rrrrrrrrrrrr|r}
\toprule
{} &  \rot{CropDisease} &  \rot{DeepWeeds} &  \rot{Flowers102} &  \rot{EuroSAT} &  \rot{Resisc45} &  \rot{ISIC} &  \rot{ChestX} &  \rot{Omniglot} &  \rot{SVHN} &  \rot{Kaokore} &  \rot{Sketch} &   \rot{DTD} &  \rot{Mean} \\
\midrule
\CE          &        96.27 &      79.93 &       87.16 &    92.58 &     82.74 & 78.16 &   45.09 &     73.39 & 59.10 &    76.13 &   66.94 & 54.41 & 74.33 \\
\MOCO        &        98.51 &      84.71 &       90.00 &    95.95 &     88.51 & 81.65 &   49.23 &     77.06 & 67.23 &    79.54 &   65.42 & 64.04 & 78.49 \\
\MIT         &        97.69 &      79.72 &       90.49 &    94.23 &     87.19 & 78.26 &   47.92 &     74.84 & 66.57 &    79.90 &   73.94 & 61.91 & 77.72 \\
\CEplusMOCO  &        98.31 &      83.75 &       91.08 &    95.31 &     88.69 & 79.89 &   48.89 &     73.52 & 61.41 &    80.88 &   72.30 & 63.30 & 78.11 \\
\MITplusMOCO &        98.37 &      83.60 &       93.14 &    95.16 &     89.66 & 80.25 &   49.14 &     75.06 & 63.29 &    81.85 &   73.85 & 64.41 & 78.98 \\
\bottomrule
\end{tabular}
    \end{adjustbox}
    \caption{Performance of different Stylized-ImageNet pretrained models on the downstream datasets in terms of  of top-1 accuracy for fixed feature linear transfer.}
    \label{tab:linear_SIN_app}
\end{table*}

\subsection{Measuring Intra-class Similarity} \label{ap:intra_class}
Supervised learning models learn feature representations by objectives that also increase the inter-class separation. However, we argue that increasing the intra-class variation, though might be harmful for in-domain performance, is beneficial for learning rich feature representations in transfer learning. We compute the inter-class and intra-class separation, as follows \cite{kornblith2020whatsinlossfunction}:
\begin{equation} \label{eq:intra_ap}
R_{\text{intra}}= \sum_{k=1}^K \sum_{i=1}^{N_k} \sum_{j=1}^{N_k} \frac{1 - \text{cosine}(\textbf{x}_{k,i}, \textbf{x}_{k,j})}  {KN_{k}^2} 
\end{equation}
\begin{equation} \label{eq:inter_ap}
    R_{\text{inter}}= \sum_{k=1}^K \sum_{\substack{1 \le m \le K\\ m \ne k}} \sum_{i=1}^{N_k} \sum_{j=1}^{N_m} \frac{1 - \text{cosine}(\textbf{x}_{k,i}, \textbf{x}_{m,j})}  {K(K-1)N_{k}^2} 
\end{equation}
where $\text{cosine}(\cdot, \cdot)$ is the cosine similarity. Table \ref{tab:R_intra_class_app} reports the intra-class and inter-class distance of the representations of penultimate layers. Surprisingly, \MIT has very high intra-class variation, although it was not trained using any such constraint. However, \CEplusMOCO does not have higher intra-class distance as we expected from MoCo loss. Our intuition is that the way we have calculated intra and inter-class distance in Eq.~\ref{eq:intra_ap} and Eq.~\ref{eq:inter_ap} might not properly capture the embedding landscape. We, nonetheless, report the scores to inform the community about our observation.  \\
\begin{table}[!htbp]
    \centering
    \begin{adjustbox}{max width=\linewidth}
    \begin{tabular}{lll}
\toprule
Method & $R_{\text{intra}}$ & $R_{\text{inter}}$ \\
\midrule
\CE       &       0.32 $\pm$ 0.0057 &       0.60 $\pm$ 0.0044 \\
\MOCO     &       0.54 $\pm$ 0.0071 &       0.77 $\pm$ 0.0036 \\
\MIT      &       0.45 $\pm$ 0.0085 &       0.71 $\pm$ 0.0055 \\
\CEplusMOCO  &       0.28 $\pm$ 0.0053 &       0.55 $\pm$ 0.0041 \\
\MITplusMOCO &       0.48 $\pm$ 0.0073 &       0.73 $\pm$ 0.0040 \\

\bottomrule
\end{tabular}
    \end{adjustbox}
    \caption{\small Intra-class and inter-class separation in the penultimate layer for different models on the ImageNet1K validation set. Results averaged over 10 runs.}
    \label{tab:R_intra_class_app}
\end{table}

\subsection{Results for CE with Stronger Augmentation} \label{ap:ce_strong_abl}

Table \ref{tab:ce_strong_ap} show comparison with \CE and \CEstrong, where \CEstrong is trained with similar data augmentation as MoCo. We note that, in linear evaluation setting, for SVHN, Sketch and Omniglot, \CEstrong performs better than \CE; however, it performs worse in CropDisease and ISIC dataset. In general, \CEstrong might be helpful for transfer learning to a domain which is very different from the source domain. We also see that contrastive approaches generally perform better in most cases, which suggests that both contrastive loss itself is helpful for learning transferable representation.   We also perform experiments where CE model is trained with similar augmentation and batch formation as in MoCo, i.e., strong augmentation and mini-batch formation with two different views of the same image, denoted as \CEmoco. Table \ref{tab:ce_aug_ap} shows results of \CEmoco. Here, we also find that it performs similar as \CE model. 

\begin{table*}[!htbp]
    \centering
    \begin{adjustbox}{max width=\linewidth}
    \begin{tabular}{llllllllllll|l}
\toprule
             \rot{CropDisease} &               \rot{DeepWeeds} &              \rot{Flowers102} &                 \rot{EuroSAT} &                \rot{Resisc45} &                    \rot{ISIC} &                  \rot{ChestX} &                \rot{Omniglot} &                    \rot{SVHN} &                 \rot{Kaokore} &                  \rot{Sketch} &                     \rot{DTD} &                    \rot{Mean} \\
\midrule
\multicolumn{13}{l}{Linear Evaluation} \\
\midrule
   95.61{\small $\pm$.08} &  83.19{\small $\pm$.26} &  85.61{\small $\pm$.09} &  94.18{\small $\pm$.23} &  84.82{\small $\pm$.05} &  76.77{\small $\pm$.45} &  43.98{\small $\pm$.33} &  66.38{\small $\pm$.42} &  70.10{\small $\pm$.06} &  72.39{\small $\pm$1.49} &  69.97{\small $\pm$.14} &  66.74{\small $\pm$.86} &  75.81{\small $\pm$.37} \\
\midrule
\multicolumn{13}{l}{Few-shot Recognition (5-shot)} \\
\midrule
       82.13{\small $\pm$.61} &  46.09{\small $\pm$.58} &  80.83{\small $\pm$.61} &  78.79{\small $\pm$.63} &  71.76{\small $\pm$.71} &  41.48{\small $\pm$.53} &  26.47{\small $\pm$.42} &  94.68{\small $\pm$.32} &  34.99{\small $\pm$.57} &  41.96{\small $\pm$.55} &  72.13{\small $\pm$.69} &  64.42{\small $\pm$.71} &  61.31{\small $\pm$.58} \\

\midrule
\multicolumn{13}{l}{Full-network finetuning} \\
\midrule
      99.90{\small $\pm$.03} &  97.17{\small $\pm$.10} &   94.30{\small $\pm$.14} &  98.65{\small $\pm$.10} &  96.00{\small $\pm$.09} &  87.34{\small $\pm$.33} &  56.01{\small $\pm$.33} &  90.64{\small $\pm$.17} &  97.14{\small $\pm$.11} &  88.12{\small $\pm$.86} &  80.56{\small $\pm$.21} &  73.42{\small $\pm$.09} &  88.27{\small $\pm$.21} \\
\bottomrule
\end{tabular}
    \end{adjustbox}
    \caption{Results with \CEstrong (CE model with strong augmentation). As mentioned in the main paper, the performance difference between \CE and \CEstrong is minor.}
    \label{tab:ce_strong_ap}
\end{table*}

\begin{table*}[!htbp]
    \centering
    \begin{adjustbox}{max width=\linewidth}
    \begin{tabular}{llllllllllll|l}
\toprule
             \rot{CropDisease} &               \rot{DeepWeeds} &              \rot{Flowers102} &                 \rot{EuroSAT} &                \rot{Resisc45} &                    \rot{ISIC} &                  \rot{ChestX} &                \rot{Omniglot} &                    \rot{SVHN} &                 \rot{Kaokore} &                  \rot{Sketch} &                     \rot{DTD} &                    \rot{Mean} \\
\midrule
\multicolumn{13}{l}{Linear Evaluation} \\
\midrule
95.49{\small $\pm$.09} &  82.21{\small $\pm$.43} &  86.94{\small $\pm$.17} &  93.79{\small $\pm$.03} &  84.29{\small $\pm$.02} &  76.86{\small $\pm$.46} &  43.82{\small $\pm$.21} &  65.95{\small $\pm$.24} &  68.61{\small $\pm$.15} &  74.75{\small $\pm$.81} &  69.39{\small $\pm$.10} &  68.00{\small $\pm$.46} &  75.84{\small $\pm$.27} \\
\midrule
\multicolumn{13}{l}{Few-shot Recognition (5-shot)} \\
\midrule
  79.47{\small $\pm$.65} &  45.03{\small $\pm$.55} &  77.92{\small $\pm$.65} &  76.54{\small $\pm$.64} &  69.62{\small $\pm$.75} &  40.61{\small $\pm$.52} &  25.44{\small $\pm$.41} &  95.56{\small $\pm$.29} &  37.68{\small $\pm$.57} &  38.39{\small $\pm$.53} &  71.26{\small $\pm$.68} &  63.04{\small $\pm$.74} &  60.05{\small $\pm$.58} \\

\midrule
\multicolumn{13}{l}{Full-network finetuning} \\
\midrule
99.89{\small $\pm$.02} &  97.31{\small $\pm$.16} &  95.03{\small $\pm$.11} &  98.87{\small $\pm$.02} &  96.09{\small $\pm$.02} &  88.38{\small $\pm$.96} &  55.93{\small $\pm$.39} &  90.39{\small $\pm$.13} &  97.21{\small $\pm$.03} &  88.15{\small $\pm$.39} &  80.33{\small $\pm$.23} &  75.59{\small $\pm$.28} &  88.60{\small $\pm$.23} \\
\bottomrule
\end{tabular}
    \end{adjustbox}
    \caption{Results with \CEmoco (CE model with similar augmentation and mini-batch formation as MoCo). We note minor performance variation between \CE and \CEmoco.}
    \label{tab:ce_aug_ap}
\end{table*}

\end{document}